\documentclass[12pt,onecolumn,draftcls]{myIEEEtran}
\usepackage{graphicx}
\usepackage{epstopdf}
\usepackage{subfigure}
\usepackage{amssymb,amsbsy,amsmath,amsfonts}
\usepackage{color}
\usepackage{multirow}
\usepackage{url}
\usepackage{array}
\usepackage{tabu}
\usepackage{algorithm}
\usepackage{algorithmic}
\usepackage[export]{adjustbox}

\title{On the Sampling Strategy for Evaluation of Spectral-spatial Methods in Hyperspectral Image Classification}

\author{Jie Liang, Jun~Zhou,~\IEEEmembership{Senior Member,~IEEE}, Yuntao~Qian,~\IEEEmembership{Member,~IEEE}, Lian Wen, Xiao Bai, and Yongsheng Gao,~\IEEEmembership{Senior Member,~IEEE}
\thanks{J. Liang is with the Research School of Engineering, Australian National University, Canberra, Australia}
\thanks{J. Zhou, L. Wen, and Y. Gao are with the Institute of Integrated and Intelligent Systems, Griffith University, Nathan, Australia. Correspondence author: J. Zhou (jun.zhou@griffith.edu.au)}
\thanks{Y. Qian is with the Institute of Artificial Intelligence, College of Computer Science, Zhejiang University, Hangzhou 310027, P.R. China. }
\thanks{X. Bai is with the School of Computer Science and Engineer, Beihang University, Beijing 100191, P.R. China}
\thanks{This research is partly supported by the National Natural Science Foundation of China projects No. 61571393.}
}

\begin{document}
\maketitle
\small
\begin{abstract}
Spectral-spatial processing has been increasingly explored in remote sensing hyperspectral image classification.
While extensive studies have focused on developing methods to improve the classification accuracy, experimental setting and design for method evaluation have drawn little attention. In the scope of supervised classification, we find that traditional experimental designs for spectral processing are often improperly used in the spectral-spatial processing context, leading to unfair or biased performance evaluation. This is especially the case when training and testing samples are randomly drawn from the same image - a practice that has been commonly adopted in the experiments. Under such setting, the dependence caused by overlap between the training and testing samples may be artificially enhanced by some spatial information processing methods such as spatial filtering and morphological operation. Such interaction between training and testing sets has violated data independence assumption that is abided by supervised learning theory and performance evaluation mechanism. Therefore, the widely adopted pixel-based random sampling strategy is not always suitable to evaluate spectral-spatial classification algorithms because it is difficult to determine whether the improvement of classification accuracy is caused by incorporating spatial information into classifier or by increasing the overlap between training and testing samples. To partially solve this problem, we propose a novel controlled random sampling strategy for spectral-spatial methods. It can greatly reduce the overlap between training and testing samples and provides more objective and accurate evaluation.
\end{abstract}


\section{Introduction}
\label{sec:intro}

Spectral-spatial processing have attracted increasing attentions during the past several years. Bringing spatial information into traditional single pixel based spectral analysis leads to better modelling of local structures in the image and facilitates more accurate land-cover and object classification.
While a large portion of the hyperspectral remote sensing community have focused their research on improving classification accuracy by developing a variety of spectral-spatial methods~\cite{Lu2007Survey, Chen2011, Fauvel2013, Moser2013Markov}, few attention has been paid to experimental settings.
Evaluation of hyperspectral image classification methods requires careful design of experiments such as appropriate benchmark data sets, sampling strategy to generate training and testing data, and appropriate and fair evaluation criteria~\cite{Lu2007Survey,Friedl2000}.
In the scope of supervised classification, we find that traditional experimental designs for spectral processing are often improperly used in the context of spectral-spatial processing, leading to unfair or biased performance evaluation. This is particularly the case when training and testing samples are randomly drawn from the same image/scene which is a common setting in the hyperspectral classification research due to limited availability of benchmark data and high cost of ground truth data collection.

\begin{figure*}[t]
\begin{center}
\includegraphics[width=0.8\textwidth]{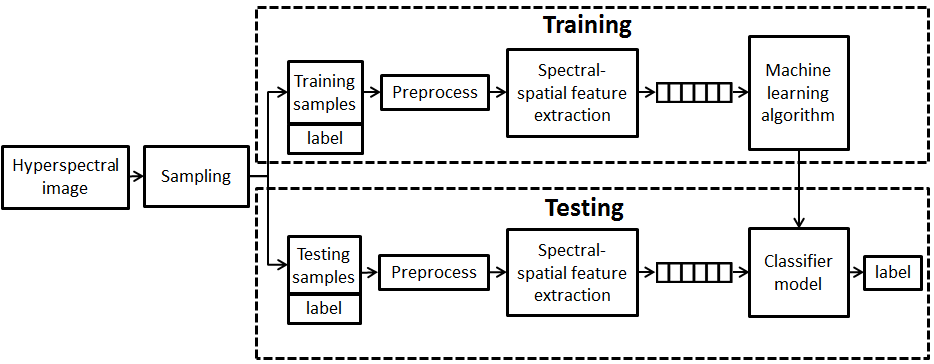}
\caption{Framework of a supervised hyperspectral image classification system that uses spectral-spatial features.}
\label{fig:framework}
\end{center}
\end{figure*}

Fig.~\ref{fig:framework} shows a typical spectral-spatial hyperspectral image classification system built on a supervised learning scheme.
Training and testing samples are drawn from an image data set following a specific sampling strategy. After image preprocessing which may involve spectral-spatial operations, feature extraction step fuses the spectral and spatial information to explore the most discriminative feature for different classes. The extracted features are used to train a classifier that minimises the error on the training set. In the testing step, the learned classifier is used to predict the classes of testing samples based on the extracted features. The testing error is given by comparing the predicted labels with the ground truth, which can be used as a performance indicator for image preprocessing, feature extraction and classification methods.

In the experimental setting, the sampling strategy plays an important role in the classifier learning and evaluation. Given a dataset including a hyperspectral image and its land-cover classes or other ground truth data, in most cases training and testing samples are not given in advance. A sampling strategy has to be employed to create the training and testing sets~\cite{Stehman1998Assessment,Richards1999,Foody2002Assessment}.
Random sampling is a natural choice since it treats all labelled data equally and each sample would be selected with the same probability.
However, by this method some classes with small number of labeled samples may have much less selected samples than expectation.
Therefore, a more sophisticated sampling method, stratified random sampling, is often used~\cite{Richards1999}. To guarantee each class having sufficient samples, it firstly groups those labelled samples into subsets based on their class labels, and then random sampling is carried out within each subset. In term of the number of training samples in each subset, it normally requires that proportion of each group should be the same as in the population. Then the rest of samples are employed as testing samples in the testing step. This method is very simple to implement, reproducible, and of statistical significance.
To the best of our knowledge, a number of hyperspectral classification methods adopted this option in the experimental setting~\cite{Chen2011,Li2012,Fauvel2013,Qian2013,Fang2014}. In the following sections, we refer to the stratified random sampling as random sampling.

Before proceeding to the issue of random sampling, we have to re-affirm some basic principles for supervised learning. Under statistical learning frame, a common assumption for inference purpose is that random variables are independent and identically distributed (\emph{i.i.d.}). The identical condition implies that training and testing samples are generated from the same data distribution. The independent condition requires that the occurrence of each sample do not affect the probability of other samples. \emph{i.i.d.} shall hold for data in different forms, for example, both raw spectral responses and extracted features. Most supervised hyperspectral image classification approaches assume that data are \emph{i.i.d.}.
Pixels in the same class shall have similar spectral responses or spectral-spatial features so that a trained classifier can be generalised to predict the labels of unseen samples. However, the independent assumption does not always hold if the training and testing samples are not carefully selected.

In general, arbitrary samples selected from a population by random sampling can be seen roughly independent from each other, or at least independent between the sets of training and testing samples. However, for hyperspectral images, the random sampling is usually undertaken on the same image. Consequently, those randomly selected training samples spread over the image and the testing samples will locate adjacent to them. Then the independence assumption would become jeopardised due to the spatial correlation between training and testing samples.
This is not a problem for the traditional pixel based spectral analysis methods in which no spatial information is used. However, when it comes to the spectral-spatial methods, the training and testing samples would inevitably interact with each other, and thus the dependence caused by overlap or partial overlap between the training and testing data could result in exaggerated classification accuracy.
To be more specific, the information from the testing set could be used in the training step by spatial operations, leading to a biased evaluation results.
The sampling problem was originally noticed by Friedl et al.~\cite{Friedl2000}, who referred to overlap as auto-correlation. Zhen et al.~\cite{Zhen2013} compared the influence of different sampling strategies to the classification accuracy. However, none of these work has given theoretical analysis on the problems and provide an effective solution. Therefore, it is necessary to revisit the sampling strategy and data dependence for supervised hyperspectral image classification, especially those based on spectral-spatial processing. In-depth discussion on this issue can be made from both experiment and the computational learning theory points of view.

In this paper, we study the relationship between sampling strategies and the spectral-spatial processing in hyperspectral image classification, when the same image is used for training and testing. We find that the experimental setting with random sampling makes data dependence on the whole image be increased by some spectral-spatial operations, and in turn increases the dependence between training and testing samples~\footnote{For the sake of conciseness and without confusion, we use ``dependence between training and testing data" and ``data dependence" interchangeably in the rest of the paper.}. To address this problem, we propose an alternative controlled random sampling strategy to alleviate the side effect of traditional random sampling on the same hyperspectral image. This leads to a fairer way to evaluate the effectiveness of spectral-spatial methods for hyperspectral classification.

In summary, the contribution of this paper are in three aspects:
\begin{itemize}
\item We point out that the traditional random sampling from the same image experimental setting is not suitable for supervised spectral-spatial classification algorithms. This helps to re-examine the performance evaluation of various spectral-spatial classification methods.
\item We find that under the random sampling setting, spectral-spatial methods can enhance the data dependence and improve the classification accuracy.
We give a theoretical explanation to this phenomenon via computational learning theory.
\item We propose a novel controlled random sampling strategy which can greatly reduce the overlap between training and testing samples caused by spatial processing, such that more objective and accurate evaluation can be achieved.
\end{itemize}

The rest of this paper is organized as follows. Section~\ref{sec:background} reviews the spectral-spatial processing that have been commonly used in hyperspectral image classification. Section~\ref{sec:spatial} provides an in-depth analysis on the dependency between training and testing samples. The spatial information embedded in the spectral-spatial processing under the experimental setting with random sampling is excavated and examined.
Section~\ref{sec:overlap} analyses the overlap between neighboring training and testing samples caused by spatial operations. Such overlap increases the dependence between training and testing samples, which may lead to mistakenly using of the testing data in the training process. Section~\ref{sec:analysis} discusses the relationship among spectral-spatial processing, data dependance and classification accuracy via computational learning theory. A new sampling strategy is proposed in Section~\ref{sec:newSampling} which reduces the influence of overlap between training and testing data.
To prove its advantage over random sampling, a series of experiments are developed and results are presented in Section~\ref{sec:ex}. At last the conclusions are drawn in Section~\ref{sec:con}.

\section{Spectral-spatial Processing in Hyperspectral Image Classification}
\label{sec:background}

The advantage of using hyperspectral data in land cover classification is that spectral responses reflect the properties of components on the ground surface~\cite{Richards1999}. Therefore, raw spectral responses can be used directly as the discriminative features of different land covers. At the same time, hyperspectral data also possesses the basic characteristic of the conventional images - the spatial information which corresponds to where a pixel locates in the image. The spatial information can be represented in different forms, such as structural information including the size and shape of objects, textures which describe the granularity and patterns, and contextual information which can express the inter-pixel dependency~\cite{Fauvel2013}.
This is also the foundation of development of most spectral-spatial methods for hyperspectral image classification.

In general, spectral-spatial information can contribute to hyperspectral image classification through three ways.
Firstly, in image preprocessing, it can be used for image denoising, morphology, and segmentation. Image denoising enables the reduction of random noises introduced from sensor, photon effects, and calibration errors. Several approaches have been exploited for this purpose, for example, smoothing filters, anisotropic diffusion, multi-linear algebra, wavelet shrinkage, and sparse coding methods~\cite{Ye2015}. In most cases, denoising can be done by applying a local filter with designed or learned kernel across the whole image. In mathematical morphology, operations are performed to extract spatial structures of objects according to their spectral responses~\cite{Velasco2010,Fauvel2013}. Similar information is explored in image segmentation, which groups spatially neighboring pixels into clusters based on their spectral distribution~\cite{Tarabalka2010,Li2012}.

Secondly, common usage of joint spectral-spatial information lies in the feature extraction stage. While traditional spectral features are extracted as responses at single pixel level in hyperspectral images, spectral-spatial feature extraction methods use spatial neighborhood to calculate features. Typical examples include texture features such as 3D discrete wavelet~\cite{Qian2013}, 3D Gabor wavelet~\cite{Jia2015}, 3D scattering wavelet\cite{Tang2015}, and local binary patterns~\cite{WLi2015}. Morphological profiles, alternatively, use closing, opening, and geodesic operators to enhance spatial structures of objects~\cite{Benediktsson2005, Fauvel2008, DallaMura2011}. Other spectral-spatial features include spectral saliency~\cite{Liang2013}, spherical harmonics~\cite{Nina2010}, and affine invariant descriptors~\cite{Khuwuthyakorn2011}. Heterogeneous features can be further fused using feature selection or reduction approaches~\cite{Jia2013}.

Thirdly, some image classification approaches rely on spatial relation between pixels for model building. A direct way of doing so is calculating the similarity between a pixel and its surrounding pixels~\cite{Pu2014}. Markov random field, for example, treats hyperspectral image as dependent data and uses spectral information in the local neighborhood to help pixel class prediction~\cite{Tarabalka2010letters,Li2012,Sun2015}. Similar spatial structures are explored in conditional random fields~\cite{Zhong2010}, hypergraph modelling~\cite{Ji2014}, and multi-scale analysis~\cite{Fang2014}. The spatial information can also be explored in constructing composite kernels in support vector machines~\cite{Camps-Valls2006}. While supervised learning approaches, such as K-nearest neighbors, linear discriminant analysis, Bayesian analysis, support vector machines, etc. are widely used in these classification tasks~\cite{Dias2013, 1993}, some approaches adopt semi-supervised or active learning strategies~\cite{Li2013,Wang2014}.

\section{Spatial Information Embedded in Random Sampling}

\label{sec:spatial}
\begin{figure*}[t]
\begin{center}
\subfigure{\includegraphics[width=0.18\textwidth, height=0.18\textwidth]{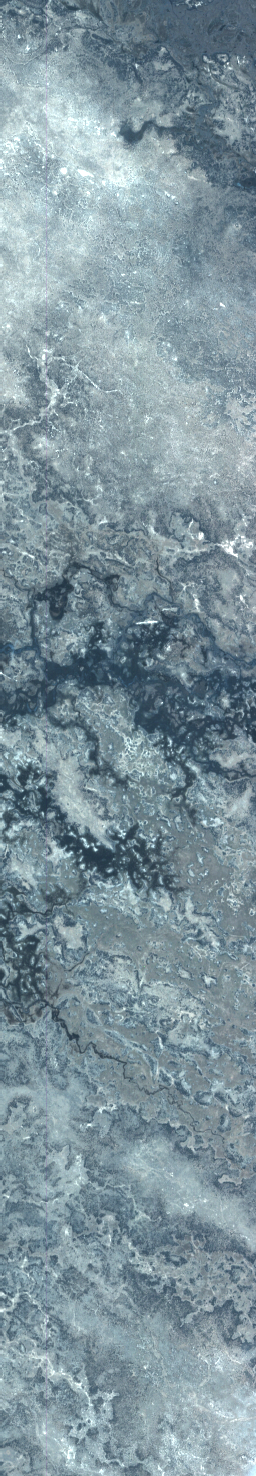}}\;
\subfigure{\includegraphics[width=0.18\textwidth, height=0.18\textwidth]{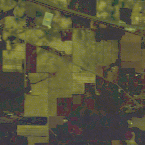}}\;
\subfigure{\includegraphics[width=0.18\textwidth, height=0.18\textwidth]{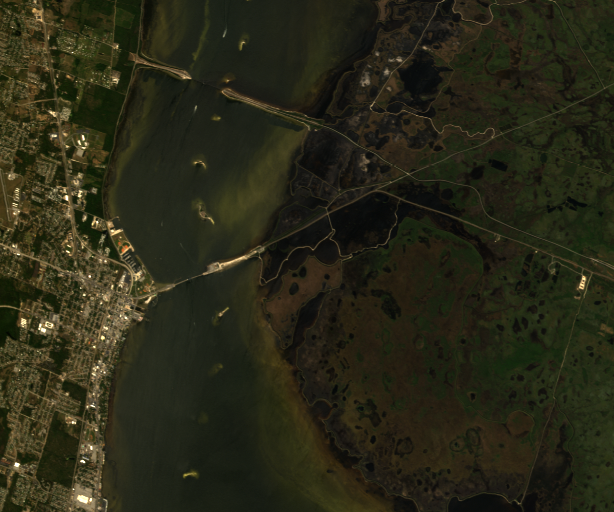}}\;
\subfigure{\includegraphics[width=0.18\textwidth, height=0.18\textwidth]{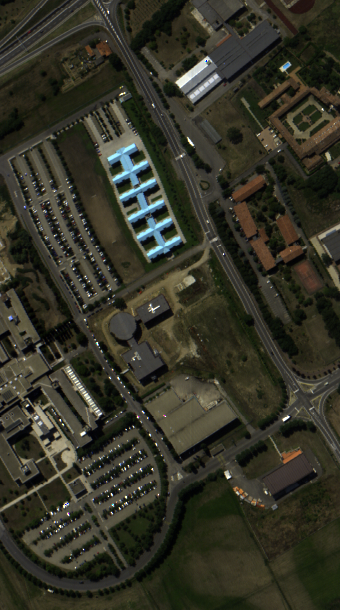}}\;
\subfigure{\includegraphics[width=0.18\textwidth, height=0.18\textwidth]{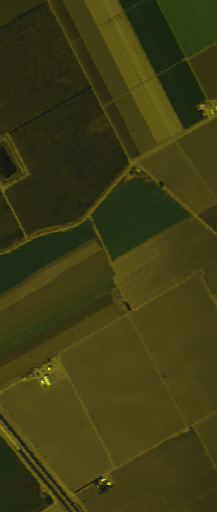}}\\

\subfigure{\includegraphics[width=0.18\textwidth, height=0.18\textwidth]{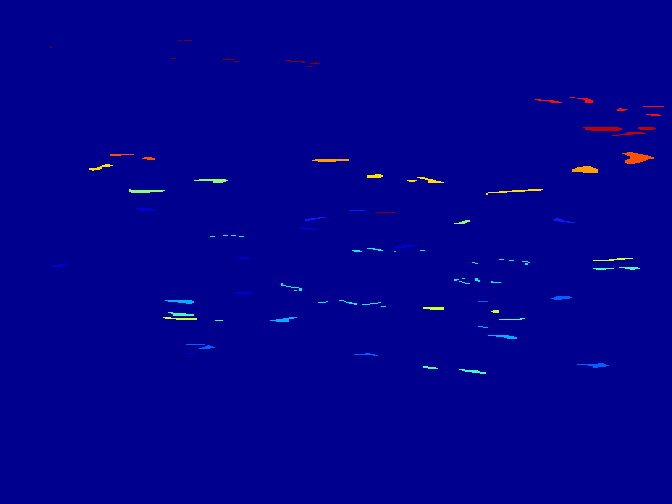}}\;
\subfigure{\includegraphics[width=0.18\textwidth, height=0.18\textwidth]{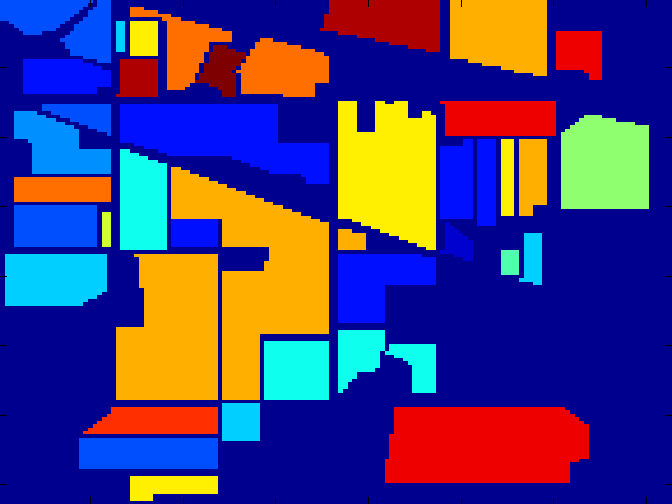}}\;
\subfigure{\includegraphics[width=0.18\textwidth, height=0.18\textwidth]{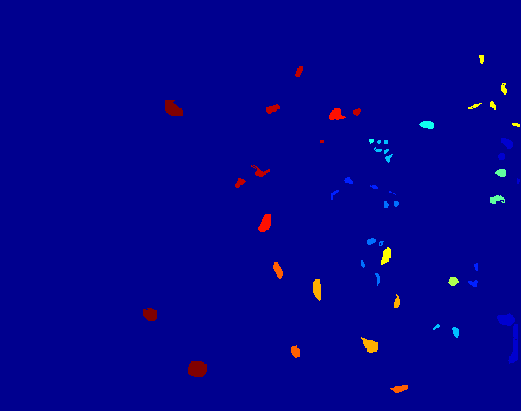}}\;
\subfigure{\includegraphics[width=0.18\textwidth, height=0.18\textwidth]{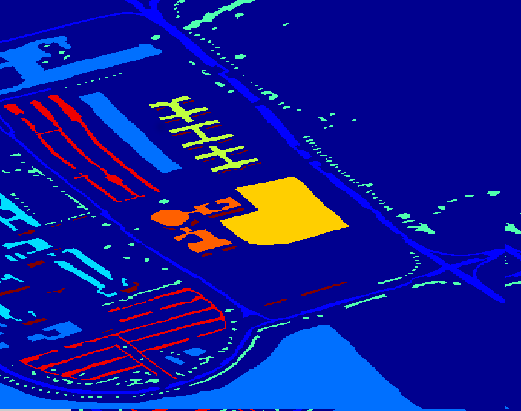}}\;
\subfigure{\includegraphics[width=0.18\textwidth, height=0.18\textwidth]{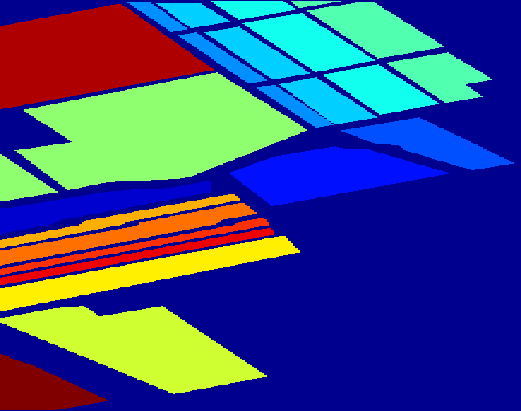}}
\caption{Three band false color composite and ground truth labels of five commonly used hyperspectral datasets. From left to right: Botswana, Indian Pines, Kennedy Space Center, Pavia University, and Salinas Scene. }
\label{fig:labels}
\end{center}
\end{figure*}

Random sampling makes the training and testing samples spread over the image, embedding plenty of underlying spatial information.
In this section, we point out that the embedded spatial information will mistakenly influence the classifier learning and evaluation.
We exploit this problem in a specific/extreme way, by which a hyperspectral classification task can even be done without spectral information.

\begin{figure}[t]
\begin{center}
\subfigure{\includegraphics[width=0.2\textwidth, height = 0.20\textwidth]{images/Indian_pines_gt.png}}\quad
\subfigure{\includegraphics[width=0.2\textwidth, height = 0.20\textwidth]{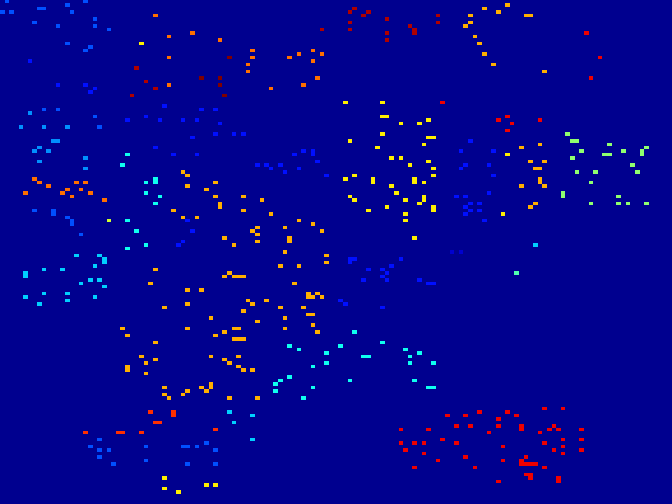}}\quad
\subfigure{\includegraphics[width=0.2\textwidth, height = 0.20\textwidth]{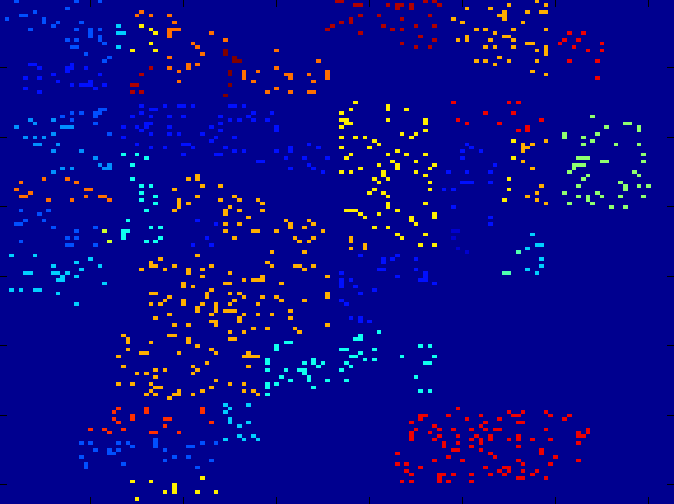}}\quad
\subfigure{\includegraphics[width=0.2\textwidth, height = 0.20\textwidth]{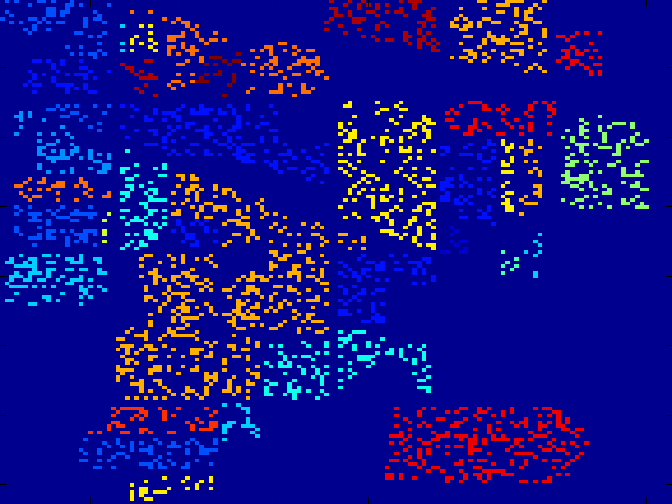}}\\
\subfigure{\includegraphics[width=0.2\textwidth, height = 0.20\textwidth]{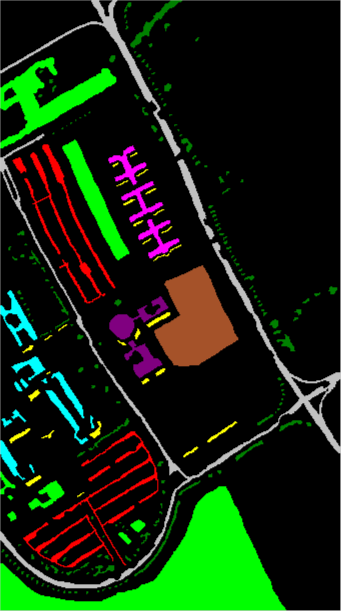}}\quad
\subfigure{\includegraphics[width=0.2\textwidth, height = 0.20\textwidth]{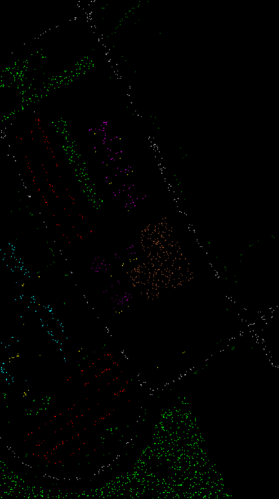}}\quad
\subfigure{\includegraphics[width=0.2\textwidth, height = 0.20\textwidth]{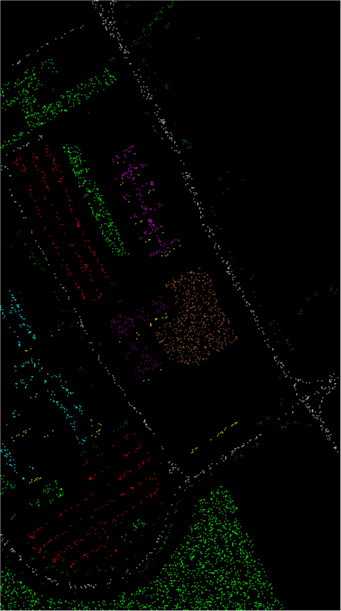}}\quad
\subfigure{\includegraphics[width=0.2\textwidth, height = 0.20\textwidth]{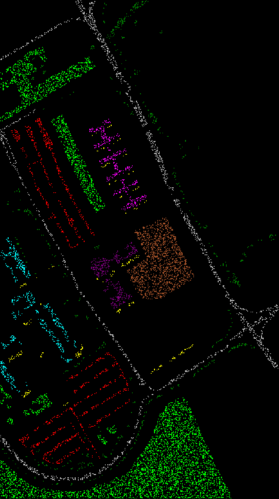}}
\caption{Random sampling strategy on Indian Pines and Pavia University datasets. From left to right: the ground truth map, training set with 5\% sampling rate, training set with 10\% sampling rate, training set with 25\% sampling rate.}
\label{fig:random}
\end{center}
\end{figure}

In many benchmark hyperspectral datasets, pixels in the same class are not distributed randomly in the image. On the contrary, they tend to exist in continuous regions and follow a certain spatial distribution, especially when objects in the same materials present in the scene. Fig.~\ref{fig:labels} shows the false color composite and ground truth maps of five commonly used hyperspectral datasets, i.e., Botswana, Indian Pines (Indian), Kennedy Space Center (KSC) , Pavia University (PaviaU), and Salinas scene (Salinas)~\cite{datasets}. In these images, there are strong dependencies between the spatial locations of pixels and land cover classes. This results in the potential using of the spatial structure and distribution of each single class.
In most cases, if random sampling is used for selecting training and testing samples in the same image, the class label of a testing sample can be easily inferred only by its spatial relation with the training samples. This can be exemplified by Fig.~\ref{fig:random}, in which 5\%, 10\% and 25\% of training data are sampled from the Indian Pines and Pavia University datasets. When it comes to 25\% sampling rate, the spatial distribution of training samples~(last column) is similar to the shape of the ground truth map~(first column) in the spatial domain.

\begin{table*}[t]
\caption{Overall accuracy (OA), average accuracy (AA) and Kappa coefficient ($\kappa$) on five hyperspectral datasets when different feature/classifier combinations were used: spectral feature with SVM~(Spe), spatial feature with SVM~(Spa1) and spatial feature with KNN~(Spa2) . }
\label{tab:comparison}
\centering
\begin{tabular}{|p{3cm}|c|c|c|c|c|c|c|c|c|}\hline
   \multirow{2}{*}{Dataset}&\multicolumn{3}{c|}{OA} & \multicolumn{3}{c|}{AA} & \multicolumn{3}{c|}{$\kappa$} \\ \cline{2-10}
                     &Spe   &Spa1   &Spa2  &Spe   &Spa1   &Spa2  &Spe   &Spa1   &Spa2 \\ \hline \hline
Botswana (\%5)	& 89.1	&\textbf{93.8}	&93.3	&	89.0	&\textbf{93.8}	&	92.9	&	0.873	&	\textbf{0.933}	&	0.928 \\ \hline
Botswana (\%10)	& 91.9	&\textbf{98.1}	&97.7	&	92.7    &\textbf{97.9}	&	97.5	&	0.913	&	\textbf{0.979}	&	0.975 \\	\hline
Botswana (\%25)	& 94.9	&\textbf{99.7}	&99.7	&	95.3	&99.6	&	\textbf{99.7}	&	0.944	&	0.996	&	\textbf{0.997} \\	\hline\hline
Indian	(\%5)	& 75.5	&\textbf{95.5}	&95.1	&	67.7	&\textbf{92.1}	&	90.5	&	0.718	&	\textbf{0.949}	&	0.944	\\	\hline
Indian	(\%10)	& 81.0	&\textbf{98.0}	&97.6	&	76.5	&\textbf{97.1}	&	94.9	&	0.783	&	\textbf{0.977}	&	0.972	\\	\hline
Indian	(\%25)	& 87.0	&\textbf{99.7}	&99.4	&	84.6	&\textbf{99.5}	&	98.7	&	0.851	&	\textbf{0.996}	&	0.993	\\	\hline\hline
KSC	(\%5)	&	87.6	&98.1	    &\textbf{98.8}	&	81.6	&	97.5	&\textbf{98.5}	&	0.862	&	0.979	&\textbf{0.987}	\\	\hline
KSC	(\%10)	&	90.3	&99.6	    &\textbf{99.8}	&	85.4	&	99.2	&\textbf{99.7}	&	0.892	&	0.995	&\textbf{0.998}	\\	\hline
KSC	(\%25)	&	93.4	&99.9	    &\textbf{100.0}	&	89.6	&	99.9	&\textbf{100.0}	&	0.927	&	1.000	&	1.000	\\	\hline\hline
PaviaU	(\%5)	& 93.2	&96.4	    &\textbf{96.9}	&	91.3	&	90.1	&\textbf{93.3}	&	0.910	&	0.952	&\textbf{0.958}	\\	\hline
PaviaU	(\%10)	& 94.2	&97.3	    &\textbf{98.7}	&	92.3	&	91.8	&\textbf{96.8}	&	0.923	&	0.964	&\textbf{0.982}	\\	\hline
PaviaU	(\%25)	& 95.3	&98.0	    &\textbf{99.7}	&	94.0	&	93.4	&\textbf{99.2}	&	0.941	&	0.973	&\textbf{0.996}	\\	\hline\hline
Salinas	(\%5)	& 93.1	&\textbf{99.9}	&99.2	&	96.2	&\textbf{99.8}	&	98.1	&	0.923	&	\textbf{0.999}	&	0.991	\\	\hline
Salinas	(\%10)	& 94.1	&\textbf{99.9}	&99.7	&	97.1	&\textbf{99.9}	&	99.4	&	0.934	&	\textbf{0.999}	&	0.997	\\	\hline
Salinas	(\%25)	& 95.3	&\textbf{100.0}	&99.9	&	97.8	&\textbf{100.0}	&	\textbf{100.0}	&	0.948	&	\textbf{1.000}	&	\textbf{1.000}	\\	\hline
\end{tabular}\end{table*}

To show the extent that the classification accuracy is impacted by spatial information, we performed experiments on five benchmark datasets in Fig.~\ref{fig:labels}. In the experiment, a nonlinear support vector machine (SVM) was employed because the land cover classes are not linearly separable in the spatial domain. The spatial coordinates were used as the spatial feature and no spectral information was included.
The parameters of the SVM were learned via five fold cross validation. Three sampling rates were explored, i.e. 5\%, 10\%, and 25\% to generate the training data from all labelled samples, while the rest of labelled data served as the testing samples.
In contrast to the spatial feature, the traditional spectral feature based methods was also implemented in which we followed the same setting as the spatial method.

Each test was repeated ten times in the experiment with random generation of training and testing samples. The overall classification accuracies (OA), average accuracies (AA) and Kappa Coefficient ($\kappa$) are shown in Table~\ref{tab:comparison} for different methods. The comparison between accuracies using spectral feature with SVM~(\textit{Spe}) and spatial feature with SVM~(\textit{Spa1}) shows some surprising results.
Classification accuracy based on pure spatial feature has significantly outperformed the counterpart using pure spectral feature in all cases. In terms of overall accuracy, the spatial method achieves more than $93.8\%$ accuracy on all datasets when only $5\%$ of training samples are used, while the spectral method has only around $75.5\%-93.2\%$ in accuracy. When the sampling rate becomes $25\%$, the accuracy almost reaches $100\%$ for the spatial feature which agrees with the perceptual intuition in Fig.~\ref{fig:random}.
Essentially, these phenomena are caused by the random sampling strategy on the same image. The results also show that higher sampling rate leads to increase of classification accuracy on all datasets.

\begin{figure}[t]
\begin{center}
\subfigure[]{\includegraphics[width=0.2\textwidth]{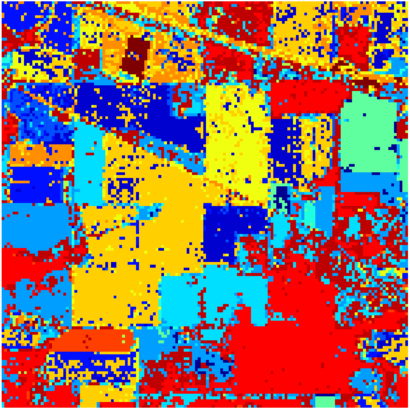}}
\subfigure[]{\includegraphics[width=0.2\textwidth]{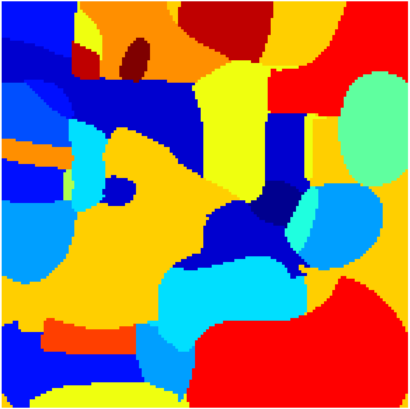}}
\subfigure[]{\includegraphics[width=0.2\textwidth]{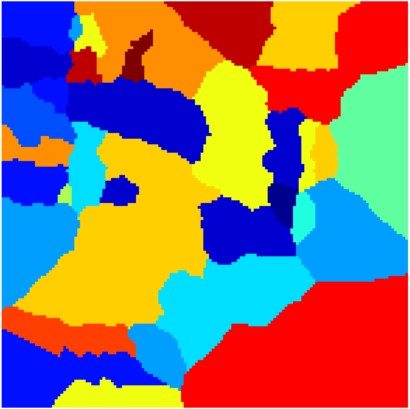}}
\caption{Classification maps of the Indian Pines~(including the unlabelled pixels) using only spectral or spatial features: (a)~\textit{Spe}, (b)~\textit{Spa1} and (c)~\textit{Spa2}.}
\label{fig:spatialMap}
\end{center}
\end{figure}

In another point of view, the spatial classification can also be exploited in the local neighbourhood. Since the training samples spread uniformly in the image, it would be easy to find a nearest training sample for any testing samples that belong to the same class. An experiment was designed to test how the local information contributes the classification. We employed the K-nearest neighbor~(KNN) classifier and set the parameter K to 1. The results are displayed in Table~\ref{tab:comparison} under the columns of \textit{Spa2}. It can be seen that the performance of \textit{Spa2} is comparable to the spatial method Spa1 on all datasets, which has significantly outperformed the spectral method on all datasets. It should be noted that in the KNN classification, predicting the label of testing samples is only based on the nearest training pixels in their spatial neighbourhood. This is similar to the mechanism of some spectral-spatial methods which also make use of the local spatial neighbourhood information but in a different way. This experiment further proves that the training data provide too much information on the spatial domain for the classification task.

While classification based on spatial coordinates seems to perform better than the spectral information, it is infeasible in real applications in which unlabelled pixels are involved. Those unlabelled pixels are prone to be classified into its nearby class, thus producing a thematic map dramatically different from the reality. To exemplify this phenomenon, Fig.~\ref{fig:spatialMap} shows the classification maps of the Indian Pines including the unlabelled pixels with 10\% sampling rate. Although \textit{Sp1} and \textit{Sp2} achieve higher classification accuracy than \textit{Spe}, their classification maps are far away from the ground truth map. Therefore this method is not acceptable in reality.
In summary, these two experiments show that random sampling from the same image makes an underestimated amount of spatial information be embedded in the training set and the testing set. It is natural to raise the concern that they would interact with each other if spatial processing is applied to the image.

\section{Overlap between Training and Testing Data from the Same Image}
\label{sec:overlap}

The spectral-spatial methods make use of the spatial information in different forms and in different ways as introduced in Section~\ref{sec:background}. When it comes to the random sampling strategy, a more severe problem may happen in the spectral-spatial analysis, especially for the feature extraction stage.
When only spectral responses are used, feature extraction is performed at single pixel, without exploring its spatial neighborhood.
Therefore, random sampling strategy provides a statistical solution for data splitting and there is no explicit overlap between training and testing samples.
However, the spectral-spatial methods usually exploit information from neighborhood pixels.
This is normally implemented by a sliding window with a specific size, for example, $3\times3$, $5\times5$ and so on.
In each window, a kernel or filter is used to extract discriminative information.
Since the training and testing samples are drawn from the same image, their features are almost certain to overlap in the spatial domain due to the shared source of information.
\begin{figure}[h]
\begin{center}
\includegraphics[width=0.45\textwidth]{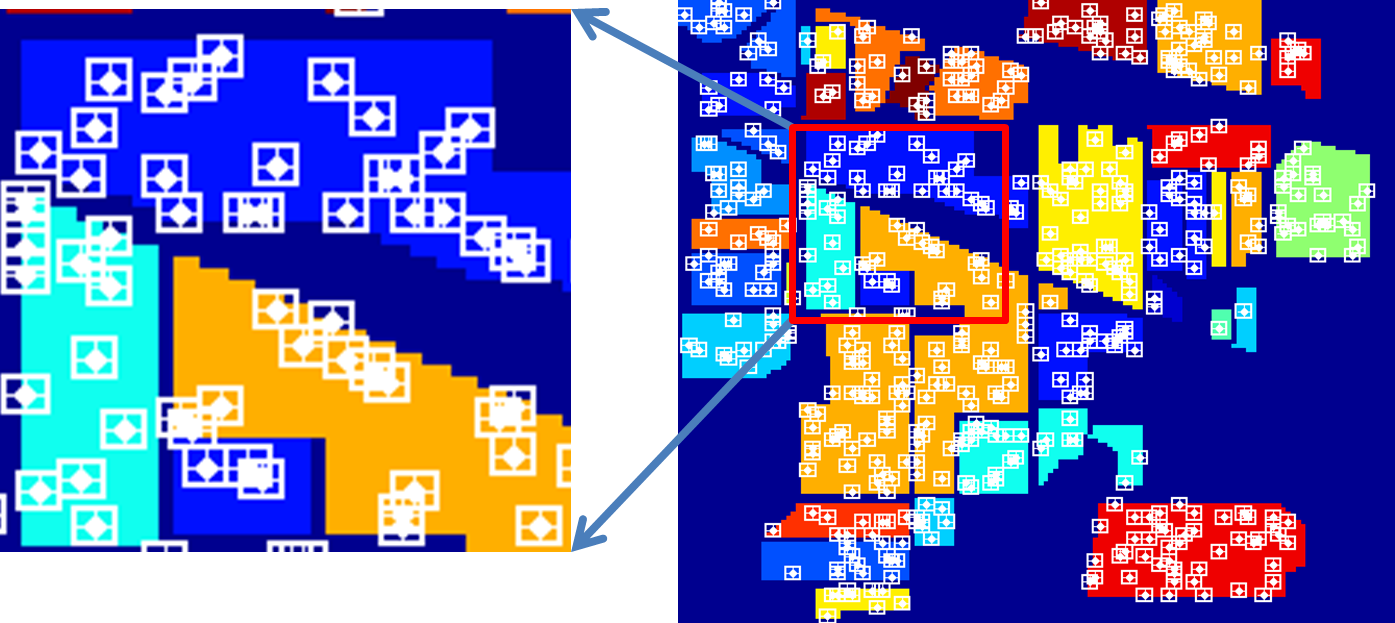}
\caption{Overlap between training and testing data on Indian Pines dataset under 5\% sampling rate. }
\label{fig:overlapExample}
\end{center}
\end{figure}
\begin{figure}[h]
\begin{center}
\subfigure[]{\includegraphics[width=0.2\textwidth]{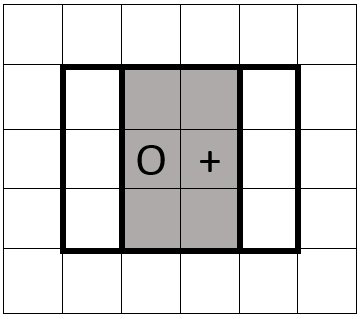}}\;
\subfigure[]{\includegraphics[width=0.198\textwidth]{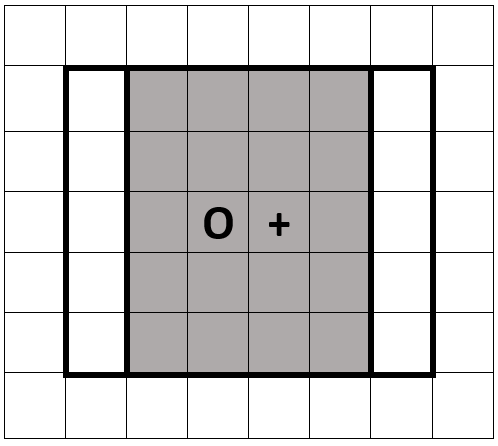}}
\caption{The regions for feature extraction from a training sample (O) and a testing sample (+) overlap with each other, as represented in gray color. The proportion of overlap is $\frac{2}{3}$ and $\frac{4}{5}$ for (a) $3 \times 3$ sliding window and (b) $5 \times 5$ sliding window, respectively.}
\label{fig:overlapdetails}
\end{center}
\end{figure}
Fig.~\ref{fig:overlapExample} shows the extent of overlap between training and testing data on the Indian Pines dataset.
In the figure, the white dots show the locations of training samples, and the surrounding white squares cover a $3\times3$ region used for spectral-spatial feature extraction. The testing samples, however, may just lie in the the square and has its own surrounding regions. This brings about a shared region between features extracted from the training and testing data such that they interact with each other and lose the mutual independence. It is also evident that a larger filter leads to more overlap areas. An example is shown in Fig.~\ref{fig:overlapdetails} in which a $3 \times 3$ and $5 \times 5$ window will result in $\frac{2}{3}$ and $\frac{4}{5}$ of overlap for adjacent training and testing samples, respectively.

Such overlap leads to using of the testing data for training purpose, and gives significant advantages to the spectral-spatial feature extraction approaches. This violates the basic principle of supervised learning that training and testing data shall not interact with each other. Depending on how feature is extracted, benefit of testing data may be explicit, for example when the spectral-spatial feature is extracted by concatenating the spectral responses of pixels in a neighborhood, or implicit, for example, by extracting texture features based on spatial frequency analysis such as discrete wavelet transform.

\subsection{Experiment with a Mean Filter Based Spectral-spatial Method}

In order to estimate how the overlap impacts the accuracy of spectral-spatial method with random sampling strategy, an experiment was carried out on the Indian Pines dataset.
In this experiment, a linear SVM classifier was used to facilitate further comparison. The features were constructed by applying a mean filter to calculate the mean of the spectral responses in a neighborhood of the hyperspectral images, which was mathematically formulated as follows:
\begin{equation}\label{eq:meanfilter}
f(x,y) =  \frac{1}{MN}\sum_{i=x-\frac{M}{2}}^{x+\frac{M}{2}}\sum_{j=y-\frac{N}{2}}^{y+\frac{N}{2}} S(i,j)
\end{equation}
where $M$ and $N$ are the width and height of neighborhood surrounding $(x,y)$. In the experiment, we set M and N both from 1 up to 27 with an interval of 2. $S(i,j)$ represents the spectral response at location $(i,j)$ and $f(x,y)$ is the feature extracted on location~$(x,y)$ which contains both spectral and spatial information. This process can be considered as one of the simplest approaches to extract spectral-spatial features.

\begin{figure}[t]
\begin{center}
\includegraphics[width=0.4\textwidth]{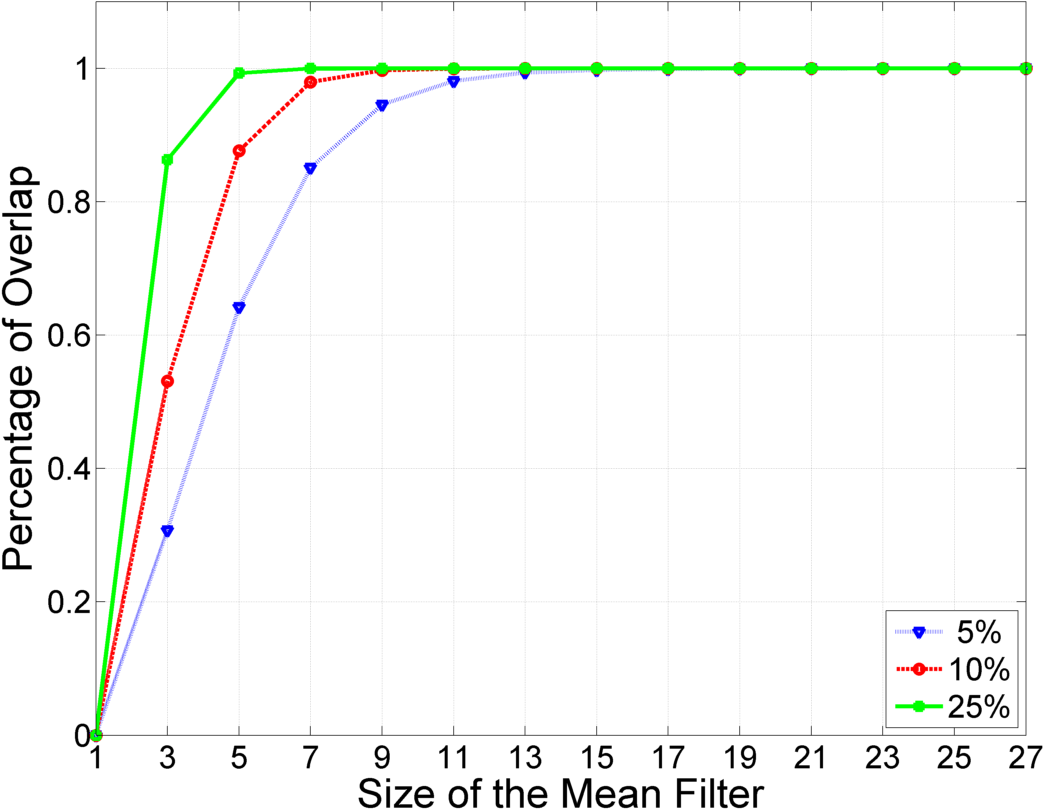}
\caption{Overlap of training and testing data on the Indian Pines with different size filters.}
\label{fig:overlap}
\end{center}
\end{figure}

When the size of the neighborhood is $1\times1$, this reduces to extracting spectral feature only. Larger size of window results in more overlap. The calculated rate of testing samples covered by the neighborhood of training samples is shown in Fig.~\ref{fig:overlap}. When 5\% training data are sampled, 30.9\% testing samples are covered by the $3\times3$ regions used to extract training features.
When random sampling rate increases to 25\%, the extent of overlap becomes 86.4\%.
The rise of sampling rate leads to rapid increase of overlap. Furthermore, when the size of filter grows, the overlap rate also increases rapidly. Eventually when the overlap rate reaches 100\%, all testing samples are used in the training process.

\begin{figure}[t]
\begin{center}
\includegraphics[width=0.4\textwidth]{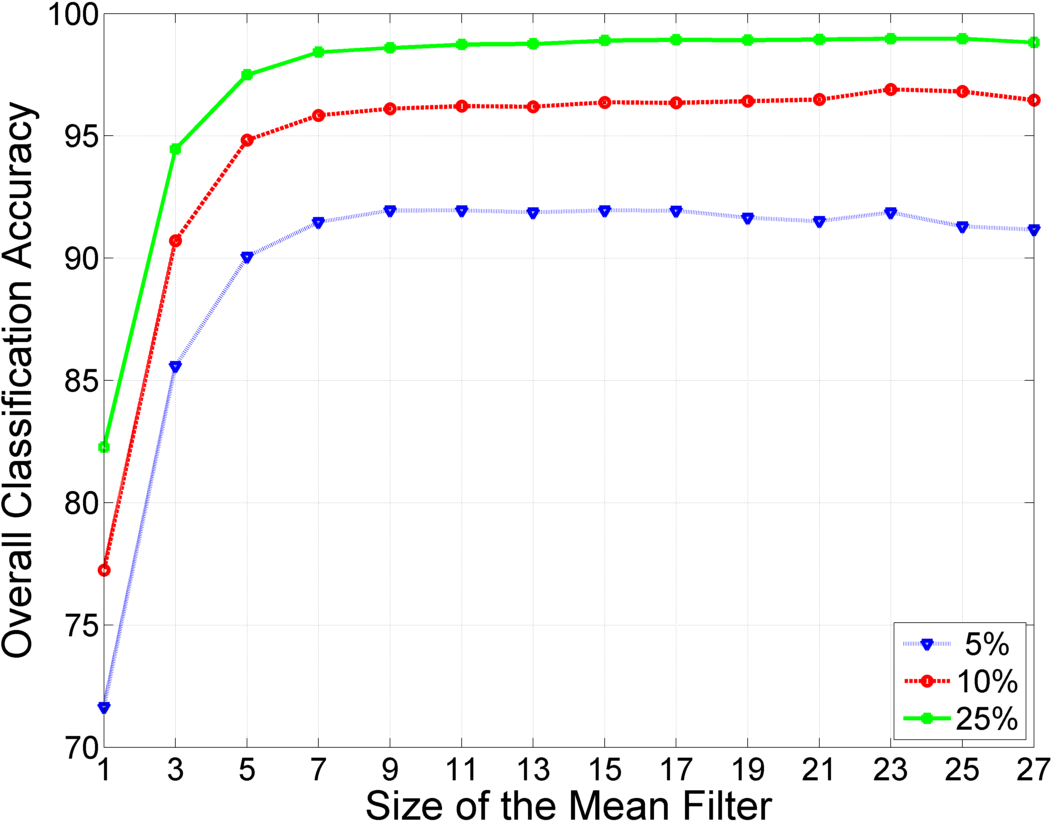}
\caption{Classification accuracies on the Indian Pines using a simple mean filter with different filter sizes. }
\label{fig:meanfilter}
\end{center}
\end{figure}

The experiment was repeated 10 times. In each time, the indices of the training and testing pixels were randomly generated. Features were generated using different settings of filter size and sampling rate. Under each setting, the same training and testing samples were used for fair comparison. The overall classification accuracies are shown in Fig.~\ref{fig:meanfilter}. Significant increase of the classification accuracy can be observed when spatial information is added to the spectral information. When the size of neighborhood increases, more testing data contribute to the training step, therefore the classification accuracy increases. It is also interesting to see that after the neighborhood increases to a specific size, the accuracy stops growing and tends to stable. This is probably because that when the neighborhood becomes too large, unlabelled data or samples from other classes are involved in the feature extraction, which neutralizes the benefits of overlap.

\subsection{Non-overlap Measurement}
Other than overlap, the increase of classification accuracy also owes to the better discriminative capability of spectral-spatial features. With larger filter size, the feature includes more spatial information. To demonstrate how the spatial neighborhood influences the effectiveness of spectral-spatial feature, we performed another experiment on those testing samples not overlapped with the training data.
\begin{table}[t]
\caption{Classification accuracies on all testing samples and non-overlapped testing samples.}
\label{tab:overlap}
\centering
\vspace{10pt}
\begin{tabular}{ |l|c|c|c|c|c|c|}
  \hline
  Filter Size     & 1             & 3              & 5          & 7     & 9    & 11        \\ \hline
  All samples (5\%)        &72.1	&86.1	&90.2	&91.4	&92.1	&92.3 \\ \hline
  Non-overlap (5\%)       &72.1	&82.9	&83.2	&79.1	&71.6	&68.0 \\ \hline
  All samples (10\%)       &77.4	&90.4	&94.5	&95.9	&96.1	&96.4\\ \hline
  Non-overlap (10\%)      &77.4	&86.2	&84.8	&77.9	&65.5	&NaN\\ \hline
  All samples (25\%)       &82.4	&94.6	&97.5	&98.3	&98.5	&98.7\\ \hline
  Non-overlap (25\%)      &82.4	&87.9	&80.6	&NaN	&NaN	&NaN\\ \hline
\end{tabular}
\end{table}

Following the same setting as the previous experiment, we removed the testing samples that were covered by the training set and only test on the remaining samples. Table~\ref{tab:overlap} shows the comparison of classification accuracy on all testing samples and non-overlap testing samples. The results show that when testing on non-overlap testing samples, the accuracy is improved when the neighbourhood information is initially introduced by the $3 \times 3$ mean filter. However, when a larger size of filter is used, the accuracy of non-overlap testing samples does not increase and even decrease~\footnote{In Table~\ref{tab:overlap}, the null values are due to the absence of non-overlapped testing samples.}. The decrease could be caused by the fact that the non-overlap testing samples are easily influenced by the samples from other classes in the neighborhood. In contrast, the classification accuracy with overlapped testing samples has remarkable improvement when larger filter size is used.

Based on the above analysis, under the random sampling strategy, some filter-based spectral-spatial feature extraction methods would make the training and testing samples overlap and then interact with each other. Subsequently, in the training process, information from testing samples are included to train the classifier, which in return is used to classify the testing samples in the testing step. Although this kind of methods improves the classification results, they are not desired because they violate the basic assumption of supervise learning and their generalization is questionable.
So far we have only analysed a special case of spectral-spatial methods, it would be interesting to extend the analysis to a broader scope. Next we try to discuss the data dependence and its impact on classification results by computational learning theory.

\section{Data Dependence and Classification Accuracy}
\label{sec:analysis}

Computational learning theory aims to analyse the computational complexity, feasibility of learning, and performance bound~\cite{Mohri2012}.
A widely known computational learning framework is the probably approximately correct~(PAC) learning which estimates the sample complexity based on the required generalization error, probability of inference and complexity of a space of functions. Another classic theory is the Vapnik-Chervonenkis theory~(VC theory). One of its functions is to bound the generalization ability of learning processes which is usually represented as the testing error $R(h)$.

Before introducing the computational learning theory, some basic learning concepts shall be firstly introduced in the scope of \emph{i.i.d.} data. In computational learning, instead of considering the classification accuracy, a more general term, generalization error bound, is usually derived to describe the ability of learning algorithm to predict the unseen data.
For a binary classification problem, given a hypothesis $h\in H$ where $H$ are all hypotheses, a target hypothesis $c$, and a sample set $S = ( x_1, x_2, ..., x_m)$ following a distribution $D$, the empirical error~(training error) $\hat{R}(h)$ and the generalization error~(testing error) $R(h)$ can be defined as:
\begin{equation}\label{eq:er}
\hat{R}(h) = \frac{1}{m}\sum_{i=1}^{m} \enspace l( h(x_i), c(x_i))
\end{equation}
\begin{equation}\label{eq:gr}
R(h) = \underset{x\in D}{\mathbb{E}} \enspace l(h(x), c(x))
\end{equation}
where $l$ is the error function and $\mathbb{E}$ is the expectation.

Despite that the empirical error $\hat{R}(h)$ can be calculated once the training data $S$, its label $c(x_i)$ and the hypothesis $h$ are known, the generalization error can not be estimated directly. In practice, simply decreasing $\hat{R}(h)$ by building complex classification model may not always minimise $R(h)$ because it may lead to over-fitting. In order to bound $R(h)$, more factors have to be considered. Based on PAC learning, the generalization bound can be calculated as:
\begin{equation}\label{eq:pac}
R(h) \leq \hat{R}(h) + \frac{1}{m}(log|H|+log\frac{1}{\delta})
\end{equation}
which means that given training data of size $m$ and hypothesis complexity $|H|$, the inequality of generalization holds with probability no less than $1-\delta$. This definition conforms to our understanding of learning that more training data leads to better learning outcome.
Based on the inequality, the generalization bound can be tightened by increasing the training sample size $m$ or by decreasing the probability $1-\delta$ which is equivalent to confidence of the inference. The complexity of hypothesis is determined by the learning models.

When the hypothesis sets are infinite, the above bound is uninformative. In order to impose generalization bound for infinite cases, the Redemacher complexity is introduced to measure the hypothesis complexity~\cite{Bartlett2002}. Specifically, it measures the variety of a set of functions by estimating the degree to which a hypothesis can fit random noise. The Rademacher complexity based generalization bound on \emph{i.i.d.} data samples is defined as:
\begin{equation}\label{eq:radiid}
R(h) \leq \hat{R}(h) + \hat{\mathfrak{R}}_s(H) + 3\sqrt{\frac{log\frac{2}{\delta}}{2m}}
\end{equation}
where $\hat{\mathfrak{R}}_s(H)$ is the empirical Rademacher complexity. $1-\delta$ is the probability or confidence and $m$ is the training sample size. $\hat{\mathfrak{R}}_s(H)$ can be estimated by growth function or VC-dimension~\cite{Mohri2012}.

Even though these models provide generalization bounds for different learning algorithms, they are all based on the \emph{i.i.d.} assumption. For \emph{non-i.i.d.} data, the generalization bound has not been fully studied due to the lack of statistical model for dependent data.
However, \emph{i.i.d.} does not always hold in practice.
In general, the samples in a hyperspectral image are not \emph{i.i.d.}, as the samples are spatially overlapping to each other in the image.
The data dependence will inevitably happen no matter how carefully the sampling strategy is designed.

In recent years, researchers begin to develop new learning theories on this topic.
Among all kinds of \emph{non-i.i.d.} data, some data types possess the property of asymptotic independence, which is weaker than independence but stronger than dependence, for instances, time series signal~\cite{McDonald2011}. In order to define this kind of data, mixing condition is used to explicitly define the dependence of the future signal on the past signal based on decay.
A commonly used model in \emph{non-i.i.d.} scenario is the stationary $\beta$-mixing model~\cite{Yu1994}. Suppose events $A$ and $B$ are generated from a time sequence $\alpha_{t\in(-\infty, +\infty)}$ with an interval $k$, the definition of $\beta$-mixing coefficient is
\begin{equation}\label{eq:beta}
\beta(k)= \sup_{m}\underset{B\in\alpha_{-\infty}^{m}}{\mathbb{E}}\left[{\sup_{A\in \alpha_{m+k}^{+\infty}}\left|Pr(A|B)- Pr(A)\right|}\right]
\end{equation}
This equation defines the dependence coefficient as the supremum of the difference between the conditional probability $Pr(A|B)$ and probability $Pr(A)$ when choosing arbitrary moment $m$ which separates event A and B.
The sequence $\alpha$ is $\beta$-mixing if $\beta(k)\rightarrow 0$ when $k\rightarrow +\infty$.
It implies that the dependence coefficient $\beta(k)$ decreases with the increase of interval $k$.

Several learning models have already been derived on stationary $\beta$-mixing data, such as VC-dimension bound\cite{Yu1994}, PAC learning~\cite{Vidyasagar2013} and Rademacher complexity~\cite{Mohri2008}. In this work, the Rademacher complexity based generalization bound is employed since it associates the generalization bounds with $\beta$-mixing coefficient. It uses a technique to transferring the original dependent data to independent blocks. Let $2\mu$ be the number of blocks and each block contains $k$ consecutive points, then the size of sample $m = 2\mu k$. The original bound in Equation~(\ref{eq:radiid}) is extended to $\beta$-mixing data as follows:
\begin{equation}\label{eq:rad}
R(h) \leq \hat{R}(h) + \hat{\mathfrak{R}}_s(H) + 3M\sqrt{\frac{log\frac{2}{\delta-4(\mu-1)\beta(k)}}{2\mu}}
\end{equation}
where $M$ is the bound of a set of hypothesis $H$.

Compared to the \emph{i.i.d.} case, this bound is not only related to the training error $\hat{R}(h)$, empirical Rademacher complexity $\hat{\mathfrak{R}}_s(H)$, and probability $\delta$, but also relies on the $\beta$-mixing coefficient $\beta(k)$ which implies the degree of dependence among data. Considering the impact of $\beta$-mixing coefficient to the bound, this equation can be further simplified as:
\begin{equation}\label{eq:sbound}
R(h) \leq f(\beta(k)) + C
\end{equation}
where $f(\beta(k))$ is a monotonically decreasing function. As a result, the generalization bound is tightened when the $\beta(k)$ increases, i.e. the dependence among data is enhanced.

Applying learning theory to hyperspectral image classification is challenging due to the complex statistical characteristic of hyperspectral images. To our knowledge, similar work in is very rare. In the following experiments, we show that hyperspectral images share the same properties of $\beta$-mixing data.

\begin{figure}[h]
\begin{center}
\subfigure[]{\includegraphics[height=0.157\textwidth]{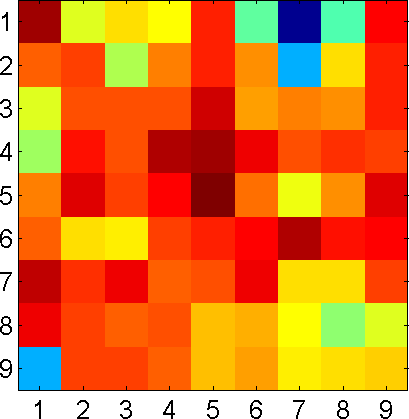}}~~~\
\subfigure[]{\includegraphics[height=0.16\textwidth]{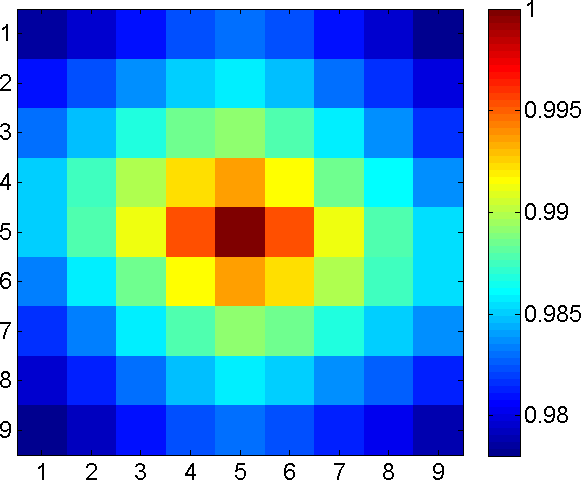}}
\caption{The correlation between a pixel and its $9\times9$ neighbourhood on Indian Pines for (a) a random location; (b) the whole image (statistical result).}
\label{fig:corr}
\end{center}
\end{figure}

Spectral feature extracted at image pixels often have strong dependence to their surrounding regions~\cite{Moser2013Markov}. However, it is still questionable whether such dependence decreases with the increasing distance between the central pixel and its neighbouring pixels. In addition, since a hyperspectral image is a three-dimensional data, how the dependence is related to the spatial direction is still unknown. To check how the dependence varies with the distance, we performed a simple statistical analysis on the Indian Pines dataset. Here, the dependence between two pixels $X$ and $Y$ is approximated by the linear correlation coefficient of their spectral responses:
\begin{equation}\label{eq:corre}
\rho_{X, Y} =\frac{cov(X,Y)}{\sigma X \sigma Y}
\end{equation}
where $cov$ and $\sigma$ represent the covariance and standard deviation, respectively.
A random location was firstly selected on this image, then the correlation coefficient $\rho$ was calculated between the pixel and its neighborhood pixels with different distances.
The result on a $9\times9$ patch is shown in Fig.~\ref{fig:corr}(a) in which the intensity implies the strength of the correlation. In the center of the patch, the intensity is 1 due to self-correlation. As expected, it does not show clear pattern at a single pixel. However, after calculating the mean of patches centered at all locations in the image, the statistical result is shown in Fig.~\ref{fig:corr}(b). It clearly shows that with the increasing interval, the correlation coefficient gradually drops in all directions. This is consistent with the characteristic of $\beta$-mixing.

\begin{figure}[t]
\begin{center}
\includegraphics[width=0.4\textwidth, height = 0.3\textwidth]{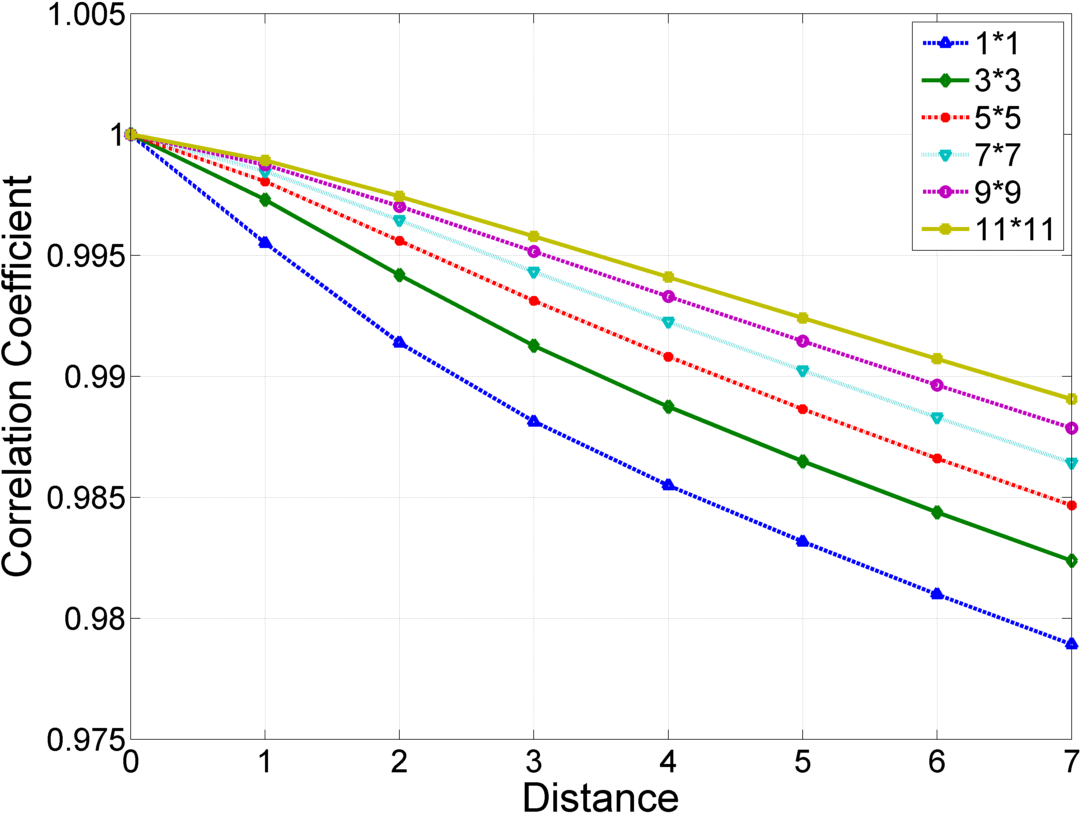}
\caption{The pixel correlation on Indian Pine processed by a mean filter with different sizes.}
\label{fig:corrfilter}
\end{center}
\end{figure}

Now we can safely assume that hyperspectral images are $\beta$-mixing, and explore the relationship between the generalization bound with data dependence. Based on Equation~(\ref{eq:sbound}), the bound is inversely related to $\beta$-mixing. As a consequence, the classification accuracy can be increased by enhancing the dependence between training and testing data. Recall that in the experiment with a mean filter based spectral-spatial method~(Fig.~\ref{fig:meanfilter}), the accuracy increases with larger filters. Similarly, the statistical results of the correlation coefficient on a $9\times9$ patch are calculated for the original image and the filtered images, which are represented by different colors. For the sake of easy observation, we only draw the distance along X-axis in the positive direction. The outcome is shown in Fig.~\ref{fig:corrfilter} from which two trends can be observed. Firstly, all curve drops continuously when the distance increases which means that the processed data agree with the properties of $\beta$-mixing. Secondly, at the same distance, the larger the filter is, the stronger the dependence between the central pixel and its adjacent pixels become. Therefore, the overlap enhances the data dependence which tighten the error bound of the final classification results.

The theories presented above have explained why mean filter improves the classification accuracy, and they can be extended to other spectral-spatial operations that increase the data dependence. It should be noted that the above analysis is built on the assumption of random sampling for performance evaluation. Under such experimental setting, the improvement of classification accuracy comes from not only incorporating spatial information into classifier but also enhancing the dependence between training and testing data. The former is the main purpose of algorithm performance evaluation and the later should be avoided.

\section{A Controlled Random Sampling Strategy}
\label{sec:newSampling}

Following the discussion in previous sections, since random sampling from the same image is not suitable for evaluation the spectral-spatial methods, it is necessary to develop a new sampling strategy to separate the training and testing sets without overlap. It would be perfect if we could perform training and testing on two different images. Unfortunately, this is still infeasible in most cases due to the limited availability of benchmark datasets and high cost of ground truth data collection. Therefore, without changing much the current experimental setting, the goal is to significantly reduce the extent of data overlap and make the evaluation fair enough.
\begin{algorithm}[t]                           
\caption{Controlled Random Sampling Strategy}    
\label{alg:controlled}                           
\begin{algorithmic}                    
    \REQUIRE Hyperspectral Image~$I$ and sampling rate~$s$
    \FOR{each class $c$ in $I$}
        \STATE Selects all unconnected partitions $P$ in the class~$c$
        \FOR{each partition $p$ in $P$}
              \STATE{Count the number of samples~$n_p$ in the partition}
              \STATE{Calculate the number of training samples~$n_t$ in the partition by $n_t = n_p \times s $ }
              \STATE{Randomly select a seed point~$q$ in the partition}
              \STATE{Applying the region-growing algorithm to extend $q$ to a region~$r$ whose size is equal to $n_t$}
        \ENDFOR
        \STATE Combine these regions~$r$ to form training samples~$R_c$
    \ENDFOR
    \STATE Combine the training samples $R_c$ and their corresponding class labels to get the whole training set $R$
\end{algorithmic}
\end{algorithm}
Based on our analysis, the main problem of random sampling is that it makes the training and testing samples spatially adjacent to each other, leading to their overlap in the subsequent spatial operations. On the other hand, as a classical method, it has advantages such as simplicity, reproducibility, and statistical significance. As a result, the new sampling strategy should satisfies the following requirements. Firstly, it shall avoid selecting samples homogeneously over the whole image so that the overlap between training and testing set can be minimised. Secondly, those selected training samples should also be representative in the spectral domain, meaning that it shall adequately cover the spectral data variation in different classes. There is a paradox between these two properties, as the spatial distribution and the spectral distribution are coupling with each other. The first property tends to make the training samples clustered so that it generates less overlap between the training and testing data. However, the second property prefers training samples being spatially distributed as random sampling does, and covering the spectral variation in different regions of the image. Therefore, a good trade-off has to be achieved by the new sampling strategy. Thirdly, because there is no prior knowledge, we do not know which samples are more important than the others. Therefore the new method shall possess the property of randomness.
\begin{figure}[t]
\begin{center}
\subfigure[Random sampling]{\includegraphics[width=0.3\textwidth, frame]{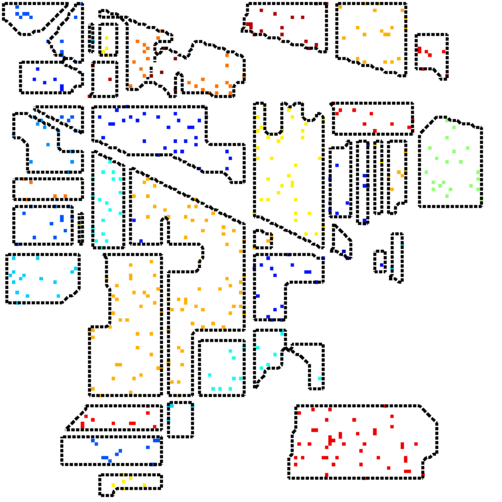}}\quad
\subfigure[Random sampling after Gaussian filtering]{\includegraphics[width=0.3\textwidth, frame]{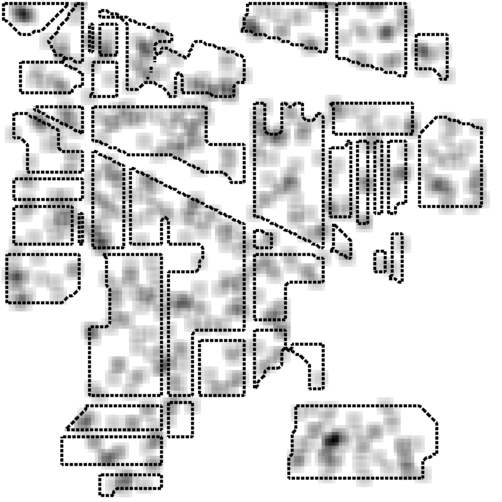}}\\
\subfigure[Controlled random sampling]{\includegraphics[width=0.3\textwidth, frame]{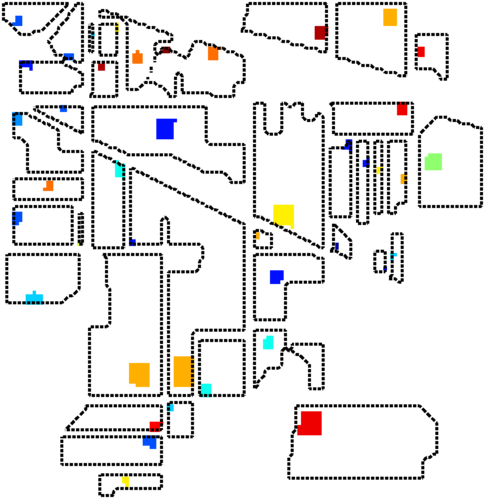}}\quad
\subfigure[Controlled random sampling after Gaussian filter]{\includegraphics[width=0.3\textwidth, frame]{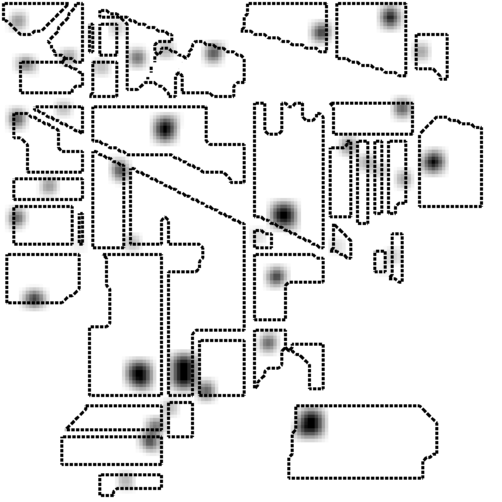}}
\caption{Overlap between the training and testing data under different sampling strategies before~(first column) and after~(second column) a Gaussian filter is applied.}
\label{fig:sketchOverlap}
\end{center}
\vspace{-10pt}
\end{figure}

Here we propose a controlled random sampling strategy to achieve a compromise of the above considerations. Similar to random sampling, a pre-defined proportion of samples in each class is to be randomly selected as the training samples and the rest data serve as the testing samples. Those training samples shall be concentrated locally and dispersed globally. We borrow the idea of region growing to create region-shape training samples~\cite{Adams1994}. The seed points are randomly selected from different partitions of classes to make the training samples disperse globally and randomly. Then controlled random sampling proceeds with three steps. Firstly, it selects the unconnected partitions for the each class and counts the samples in each partition.
This step is to find the spatial distribution of each class and make sure that the selected training samples in the next step cover the spectral variance at the most extent. Secondly, for each partition, the training samples are generated by extending region from the seed pixel. In terms of region growing, it expands in all directions and take account of 8-connected neighborhood pixels.
All the adjacent pixels of seed pixels are examined and if they are within the same class, they work as the new seed points. This process is repeated until the amount of selected points reach a pre-defined number which is proportional to the number of pixels in the corresponding partition. This guarantees that the total number of training samples meet the pre-defined proportion of the whole data population.
Thirdly, after the above steps are applied to all classes, those samples in the grown regions with their labels are chosen as the training samples and the rest of pixels work as the testing samples. In case when there are more partitions than the required training samples, partitions are again randomly sampled. A summary of this strategy is given in Algorithm~\ref{alg:controlled}.
\begin{figure}[t]
\begin{center}
\subfigure{\includegraphics[width=0.2\textwidth,height = 0.20\textwidth]{images/Indian_pines_gt.png}}\quad
\subfigure{\includegraphics[width=0.2\textwidth,height = 0.20\textwidth]{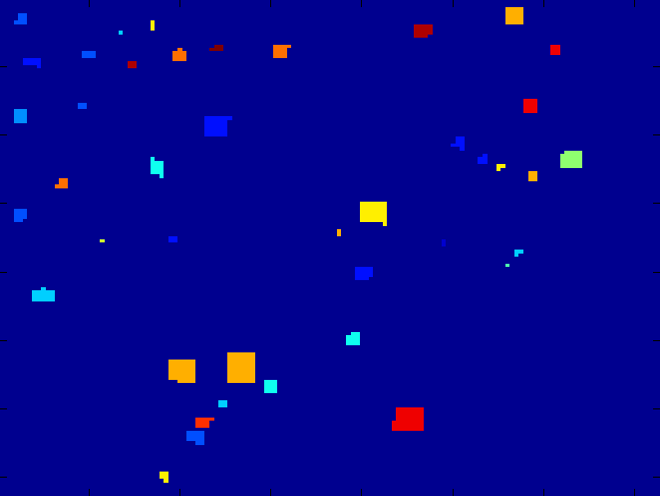}}\quad
\subfigure{\includegraphics[width=0.2\textwidth,height = 0.20\textwidth]{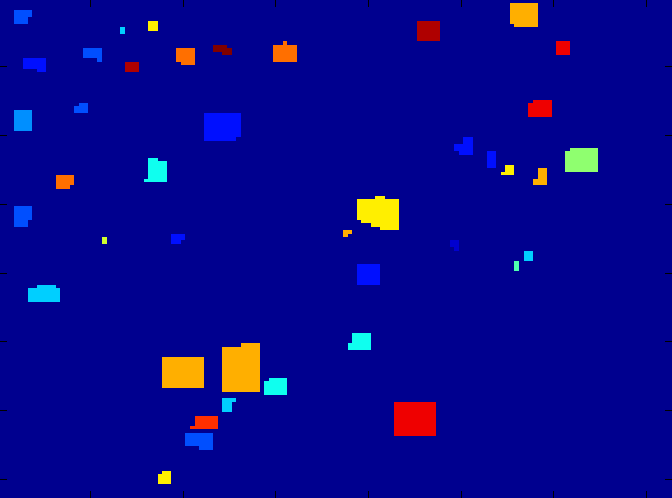}}\quad
\subfigure{\includegraphics[width=0.2\textwidth,height = 0.20\textwidth]{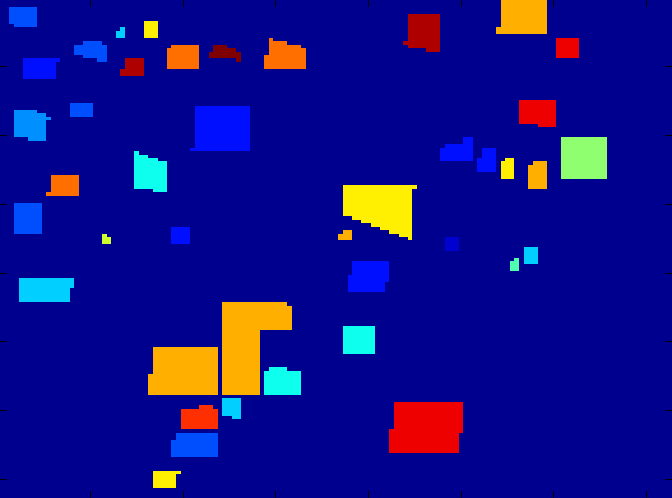}}\\
\subfigure{\includegraphics[width=0.2\textwidth,height = 0.20\textwidth]{images/PaviaU_groundTruth.png}}\quad
\subfigure{\includegraphics[width=0.2\textwidth,height = 0.20\textwidth]{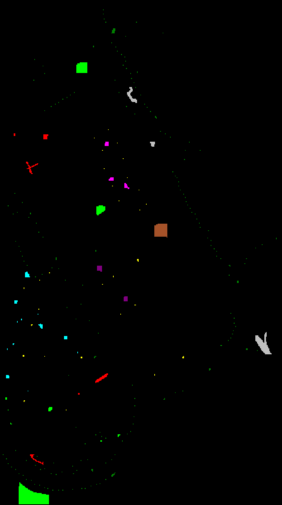}}\quad
\subfigure{\includegraphics[width=0.2\textwidth,height = 0.20\textwidth]{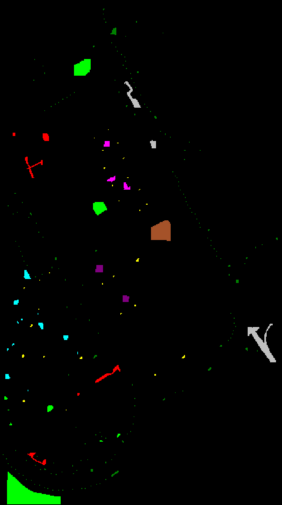}}\quad
\subfigure{\includegraphics[width=0.2\textwidth,height = 0.20\textwidth]{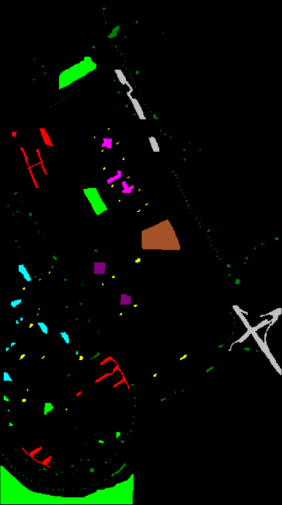}}
\caption{Controlled random sampling strategy on the Indian Pines and Pavia University datasets. From left to right: the ground truth map, training set with 5\%, 10\%, and 25\% sampling rates, respectively.}
\label{fig:randomRegion}
\end{center}
\vspace{-10pt}
\end{figure}

In Fig.~\ref{fig:sketchOverlap}, we demonstrate different degrees of overlap between training and testing samples under random sampling and controlled random sampling strategies, after a Gaussian filter is applied. In the left column of the figure, the training and testing data are represented by colored dots and white regions in each partition. Applying the Gaussian filter creates the gray regions in the right column of the figure, representing the overlap between the training and testing data. It can be noticed that all the training samples are impacted by the testing data under random sampling. On the contrary, for controlled random sampling, only training samples at the edges of the training regions are influenced by the testing data.
This figure clearly shows that the overlap from controlled random sampling is significantly less than that from the traditional random sampling.

To further illustrate how the controlled random sampling works with real datasets, examples on Indian Pines and Pavia University are given in Fig.~\ref{fig:randomRegion} with 5\%, 10\% and 25\% sampling rates. Compared to the random sampling strategy in Fig.~\ref{fig:random}, it can be observed that the spatial structure of each class can no longer be inferred from the training data as random sampling does. In the meantime, the training samples are still distributed across the whole image and a wide range of spectral variances are covered. Though this approach can not completely eliminate overlap between the training and testing data, the influence of testing data in the training stage can be greatly reduced to limited pixels at the boundaries of each training region. The experimental setting with the proposed sampling method can help us more accurately and objectively evaluate the performance of spectral-spatial methods.

\section{Experiments}
\label{sec:ex}
To prove the usability and advantage of the proposed controlled random sampling against random sampling, we have developed a series of experiments to test these two strategies when they are used to evaluate spectral-spatial operations in different stages of image classification. In the preprocessing step, we adopted a mean filter and a Gaussian filter as examples of smoothing and denoising operations. Then, we performed experiments with raw spectral feature to examine the effectiveness of the proposed sampling method when evaluating the spectral responses without spatial processing.
Finally, two spectral-spatial feature extraction methods, i.e. 3D discrete wavelet and morphological profiles, were compared using two sampling methods. In order to make the experiments more convincing, we adopted two widely used supervised classifiers, support vector machine~(SVM) and random forest~(RF)~\cite{Gislason2006} to validate our results. The SVM was implemented using the LIBSVM package~\cite{Chang2011}, and the RF was implemented using the well-known Weka 3 data mining toolbox~\cite{Hall2009}. We present results on five benchmark datasets, i.e., Botswana, Indian Pines (Indian), Kennedy Space Center (KSC), Pavia University (PaviaU), and Salinas scene (Salinas).

\subsection{Evaluation of Spectral-spatial Preprocessing Method}
\begin{figure}[t]
\begin{center}
\subfigure[Indian Pines \& SVM]{\includegraphics[width=0.23\textwidth, height = 0.2\textwidth]{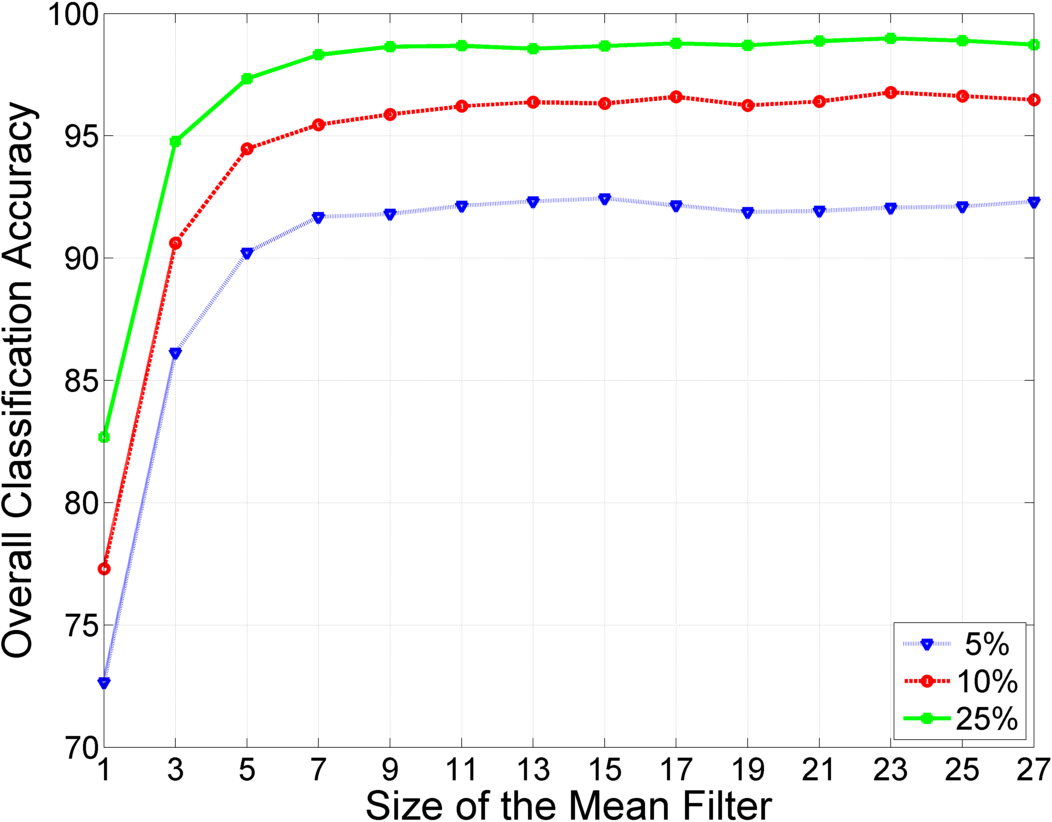}}
\subfigure[Indian Pines \& RF]{\includegraphics[width=0.23\textwidth, height = 0.2\textwidth]{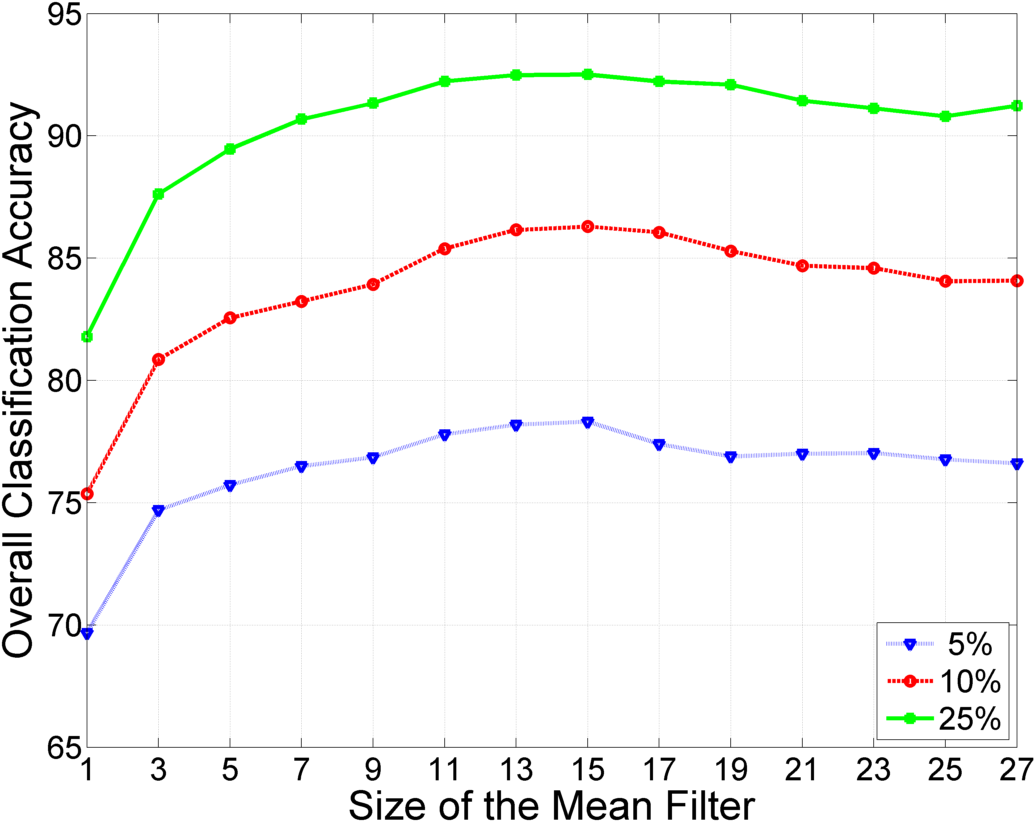}}
\subfigure[PaviaU \& SVM]{\includegraphics[width=0.23\textwidth, height = 0.2\textwidth]{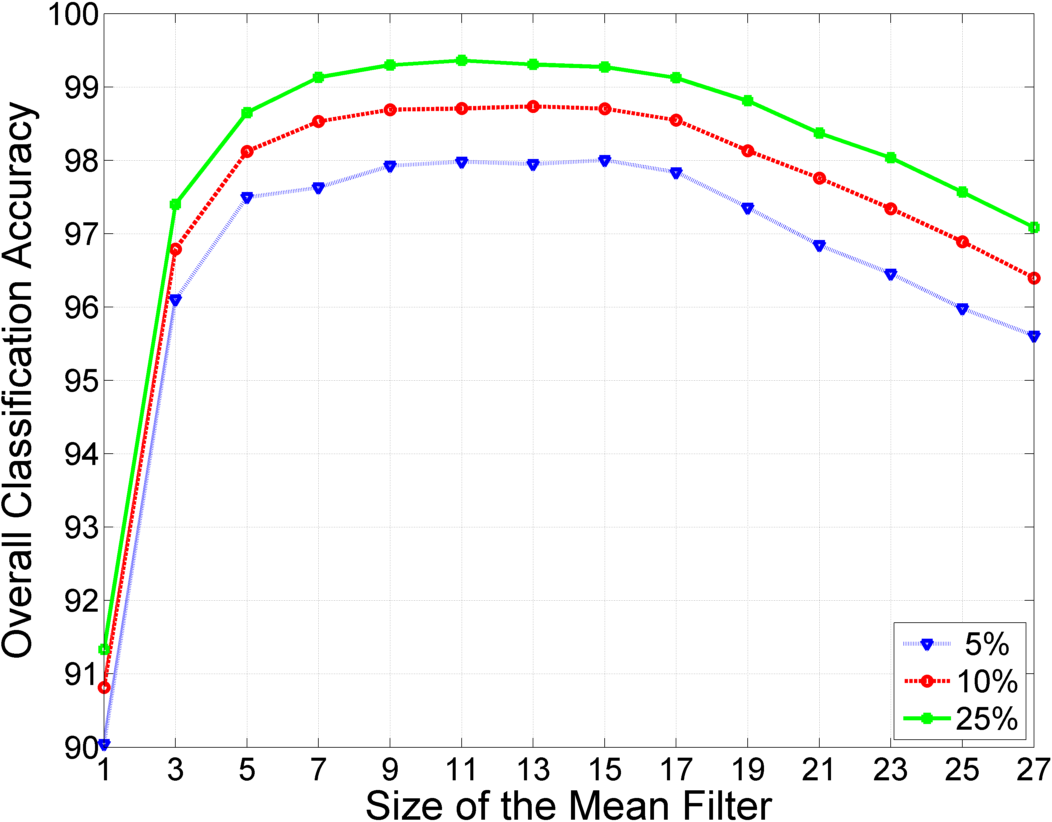}}
\subfigure[PaviaU \& RF]{\includegraphics[width=0.23\textwidth, height = 0.2\textwidth]{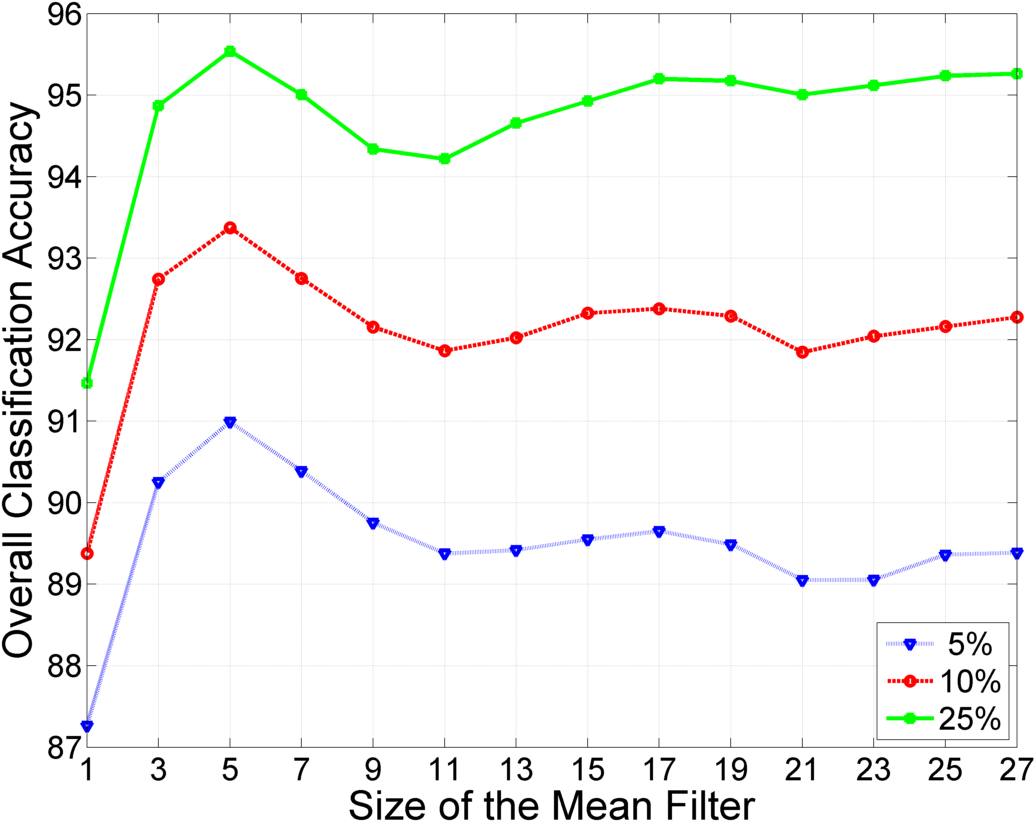}}
\\
\subfigure[Indian Pines \& SVM]{\includegraphics[width=0.23\textwidth, height = 0.2\textwidth]{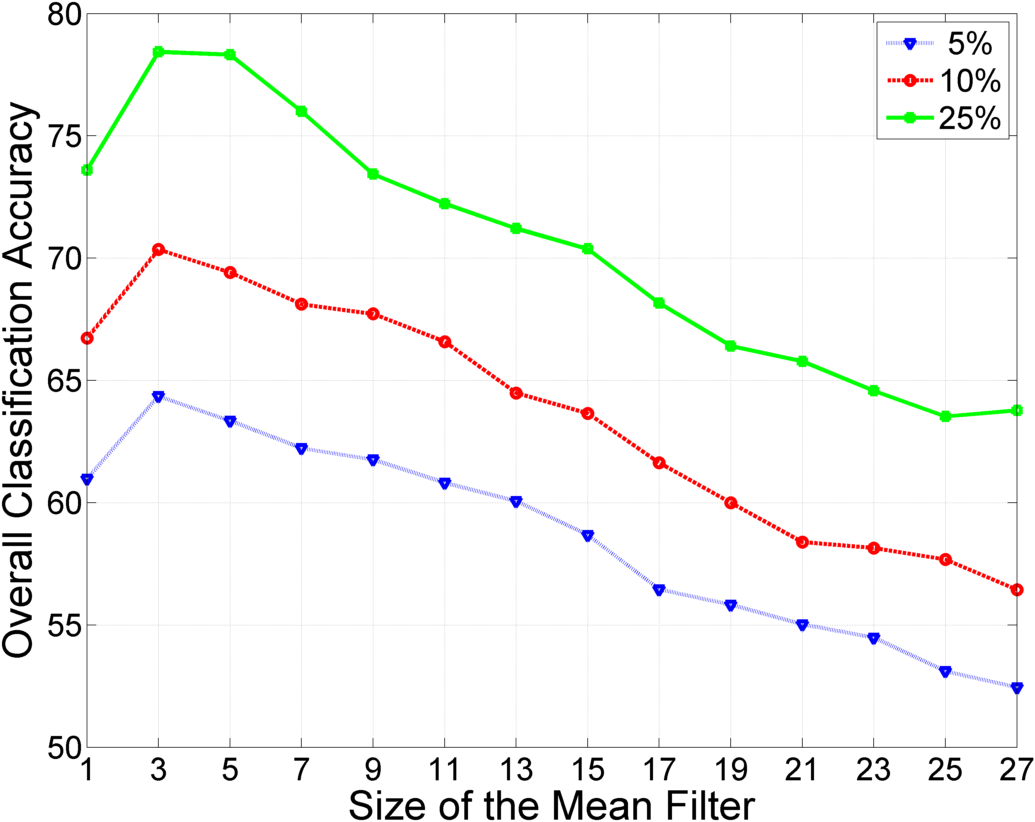}}
\subfigure[Indian Pines \& RF]{\includegraphics[width=0.23\textwidth, height = 0.2\textwidth]{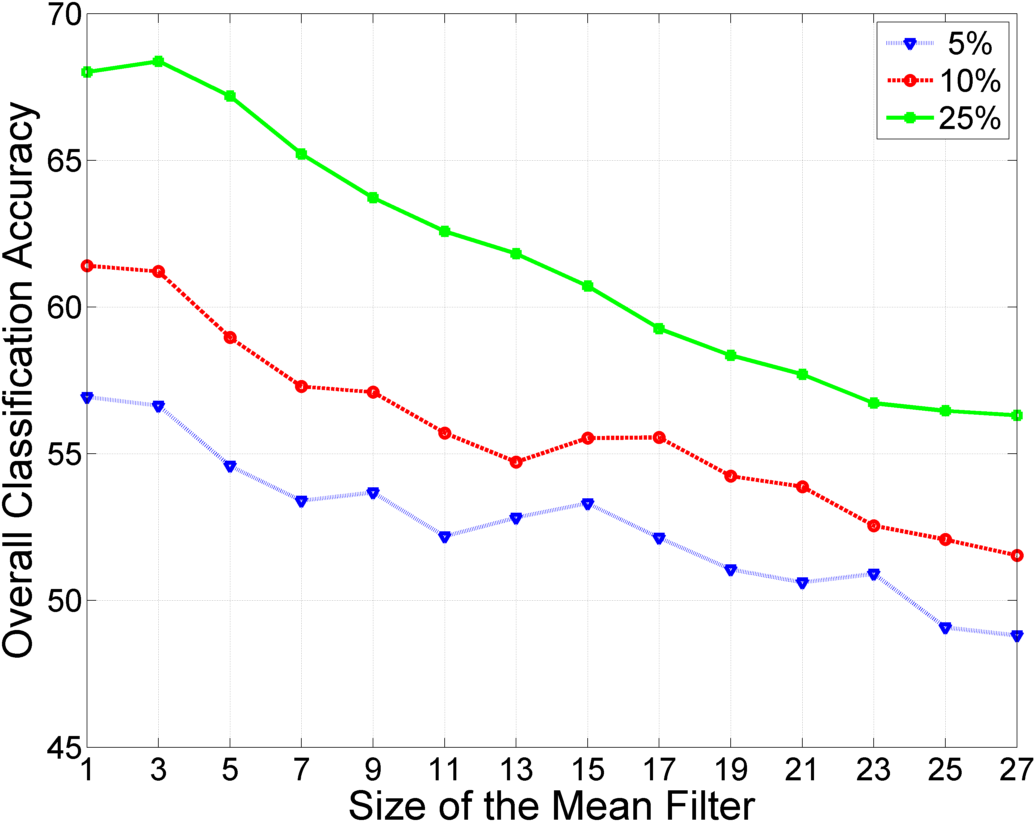}}
\subfigure[PaviaU \& SVM]{\includegraphics[width=0.23\textwidth, height = 0.2\textwidth]{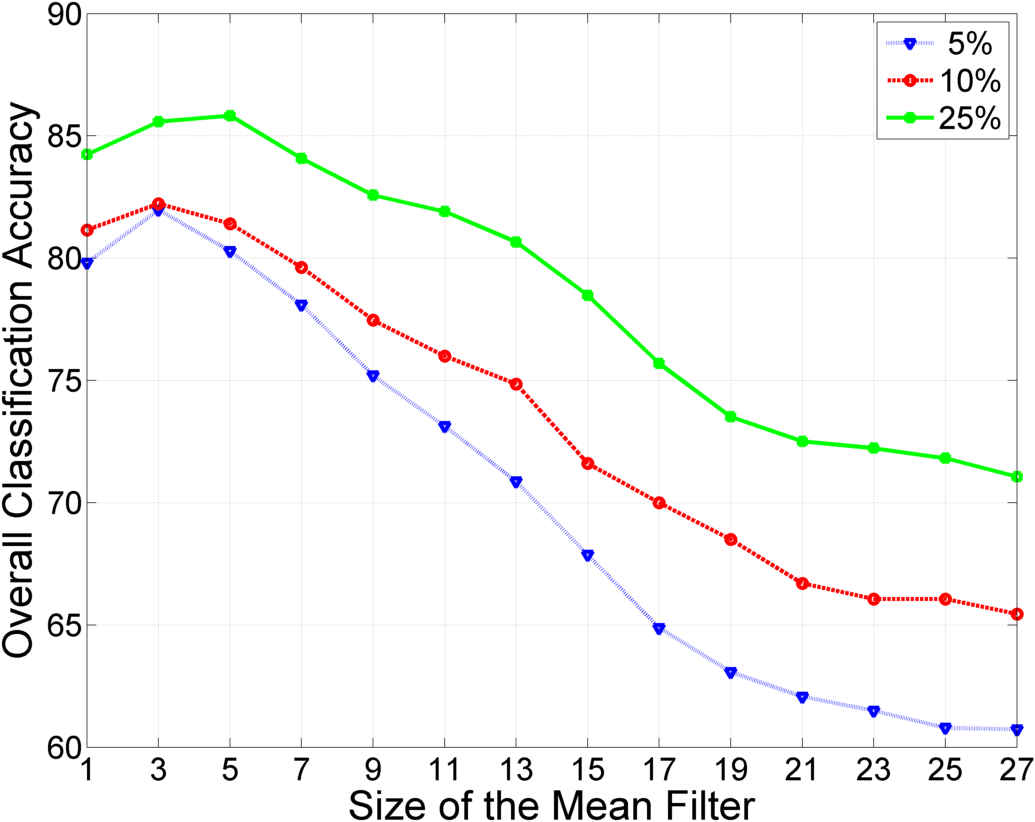}}
\subfigure[PaviaU \& RF]{\includegraphics[width=0.23\textwidth, height = 0.2\textwidth]{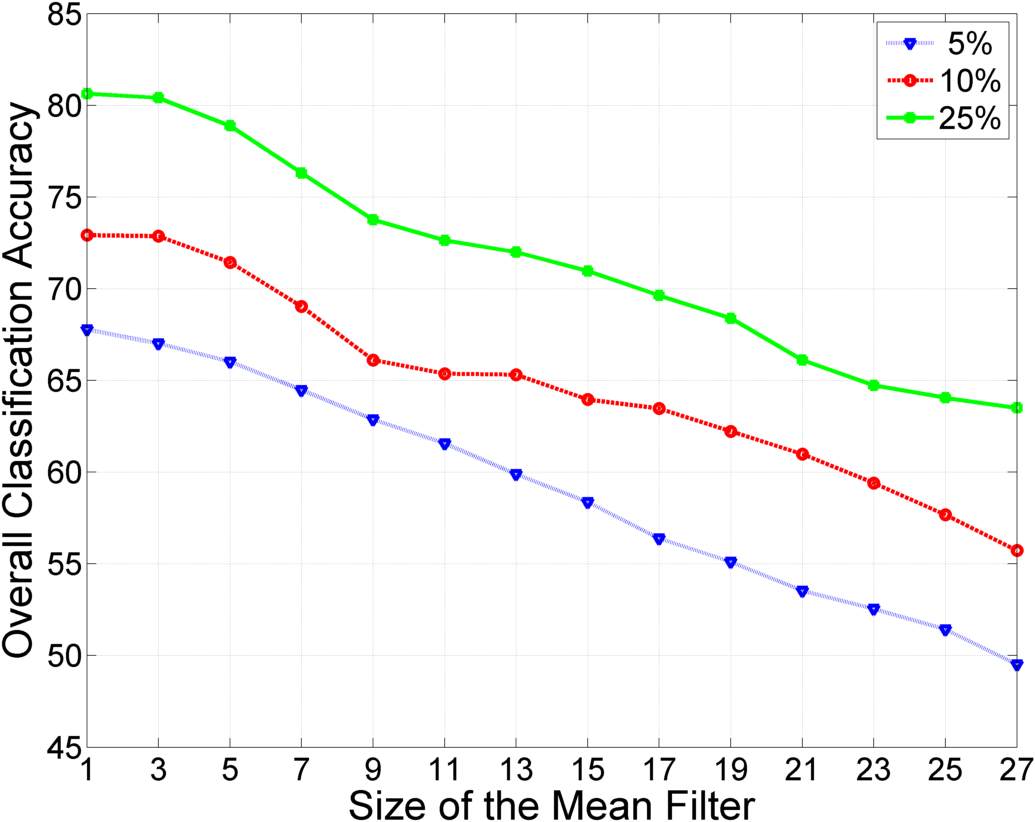}}
\caption{Classification accuracies vary with the size of mean filter on the Indian Pines and Pavia University~(PaviaU) datasets under random sampling~(first row) and controlled random sampling~(second row) strategies.}
\label{fig:preprocessing1}
\end{center}
\end{figure}

The spectral-spatial preprocessing step contributes to classification by improving the quality of hyperspectral images, reducing random noises, and enhancing specific features. By varying the parameters of mean filter and Gaussian filter, their influence to the classification accuracy under two sampling strategies can be analysed. We undertook experiments on both Indian Pines and Pavia University datasets with SVM and RF, respectively. The results with mean filter are shown in Fig.~\ref{fig:preprocessing1}. When traditional random sampling is used, the accuracy on the Indian Pines dataset increases with larger filter size when SVM and RF are adopted~(Fig.~\ref{fig:preprocessing1}(a) and (b)). For the Pavia University dataset~(Fig.~\ref{fig:preprocessing1}(c) and (d)), the accuracies also increase with larger filter size but decrease when the size reaches a specific value, which is slightly different from the results on the Indian Pines image. The reason may be that Pavia University has higher spatial resolution and interacts with filters in more complex way than the low spatial resolution Indian Pines data. The results confirm that using a mean filter with relative large size can increase the classification accuracy, up to 92.4\% on Indian Pines and 98.0\% on Pavia University. Essentially, it is mainly because larger filter leads to more overlap between the training and testing data. In contrast, when adopting the controlled random sampling strategy, the classification accuracy firstly improves marginally, but then dramatically drops with larger size filters. This is consistent with our expectation in evaluating the influence of spectral-spatial operations rather than the data dependence. Therefore, the proposed sampling method successfully avoids the problem of random sampling.

We then performed an experiment with Gaussian filter under the same setting to compare two sampling strategies. Among different denoising and smoothing approaches, Gaussian filter is a basic but effective tool to reduce the random noise in hyperspectral images. It works as a low-pass filter whose standard deviation controls the shape of filter and sets the threshold to remove the corresponding high frequency signal. The larger the stand deviation is, the lower frequency the signal can be preserved and the image be more smoothed. We applied a Gaussian filter on each band of hyperspectral images with a range of standard deviations. The size of filters varies with the standard deviation so that the smoothing effect decays to nearly zero at the boundaries of filtering masks. Then the smoothed image was fed into the classifier. This experiment was repeated 10 times and the mean of overall accuracy was used as the evaluation criterion. The standard deviation ranged from $2^{-1}$ to $2^3$ with an interval of 0.5 on the exponential term.

\begin{figure}[t]
\begin{center}
\subfigure[Indian Pines \& SVM]{\includegraphics[width=0.23\textwidth, height = 0.2\textwidth]{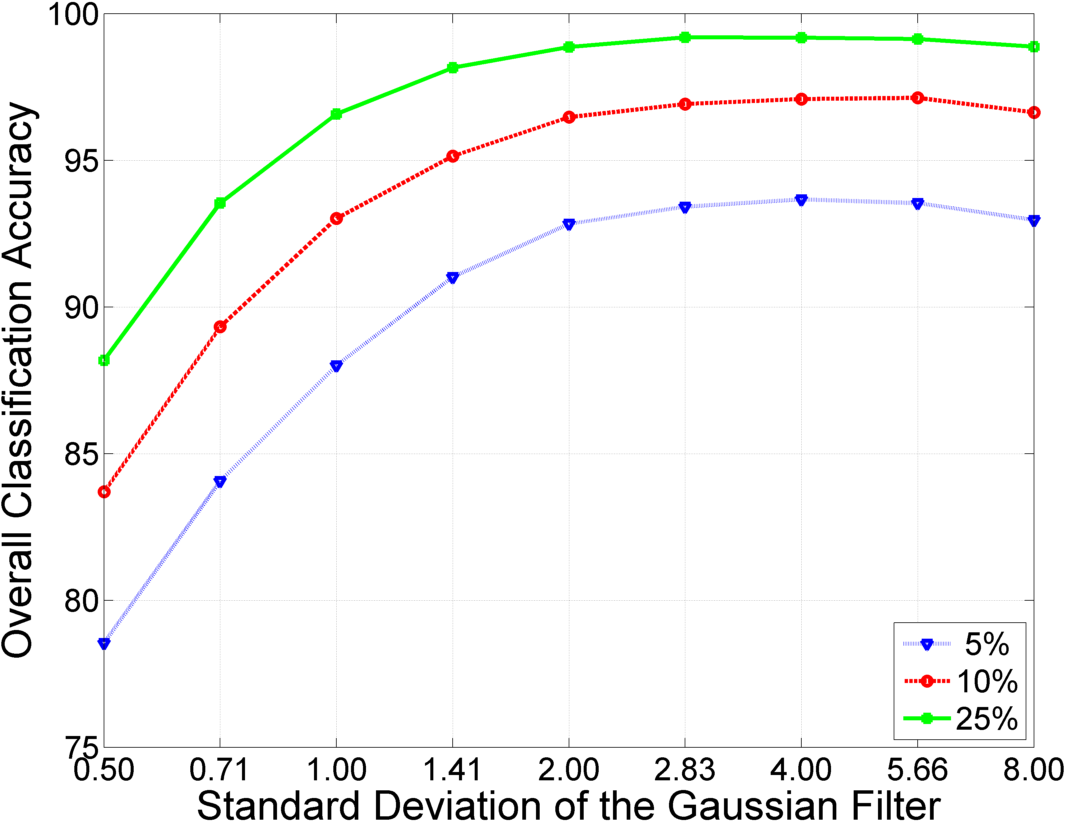}}
\subfigure[Indian Pines \& RF]{\includegraphics[width=0.23\textwidth, height = 0.2\textwidth]{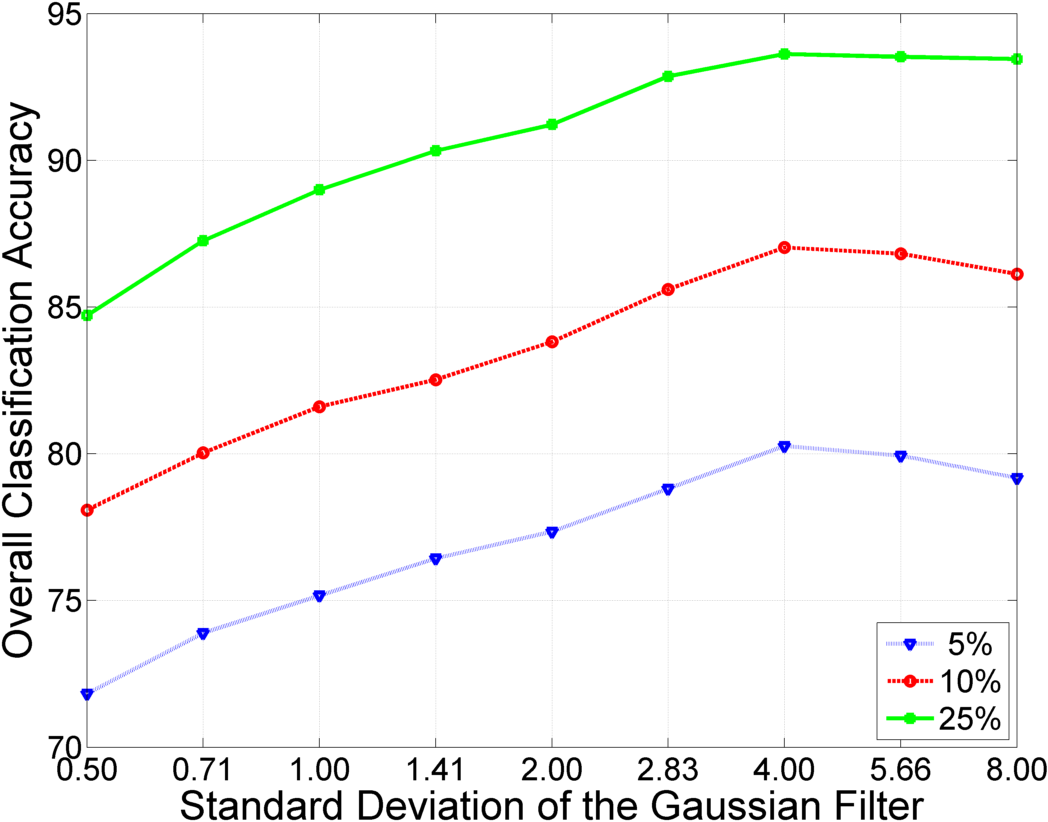}}
\subfigure[PaviaU \& SVM]{\includegraphics[width=0.23\textwidth, height = 0.2\textwidth]{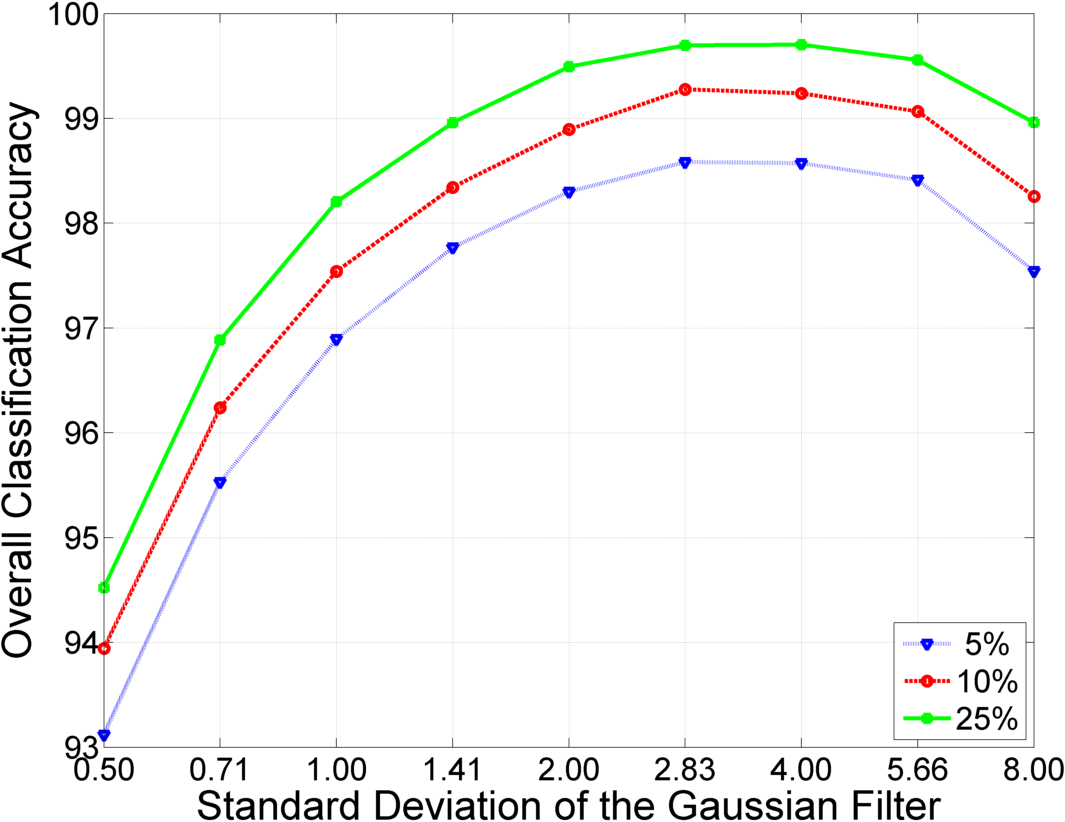}}
\subfigure[PaviaU \& RF]{\includegraphics[width=0.23\textwidth, height = 0.2\textwidth]{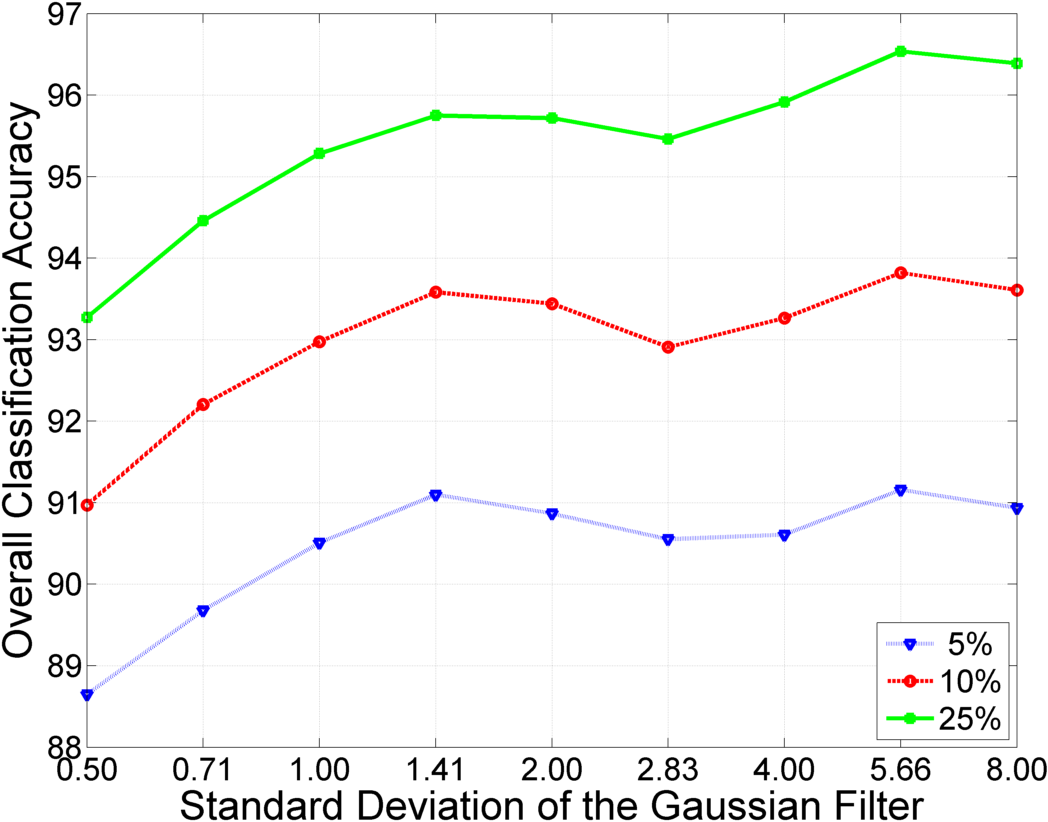}}
\\
\subfigure[Indian Pines \& SVM]{\includegraphics[width=0.23\textwidth, height = 0.2\textwidth]{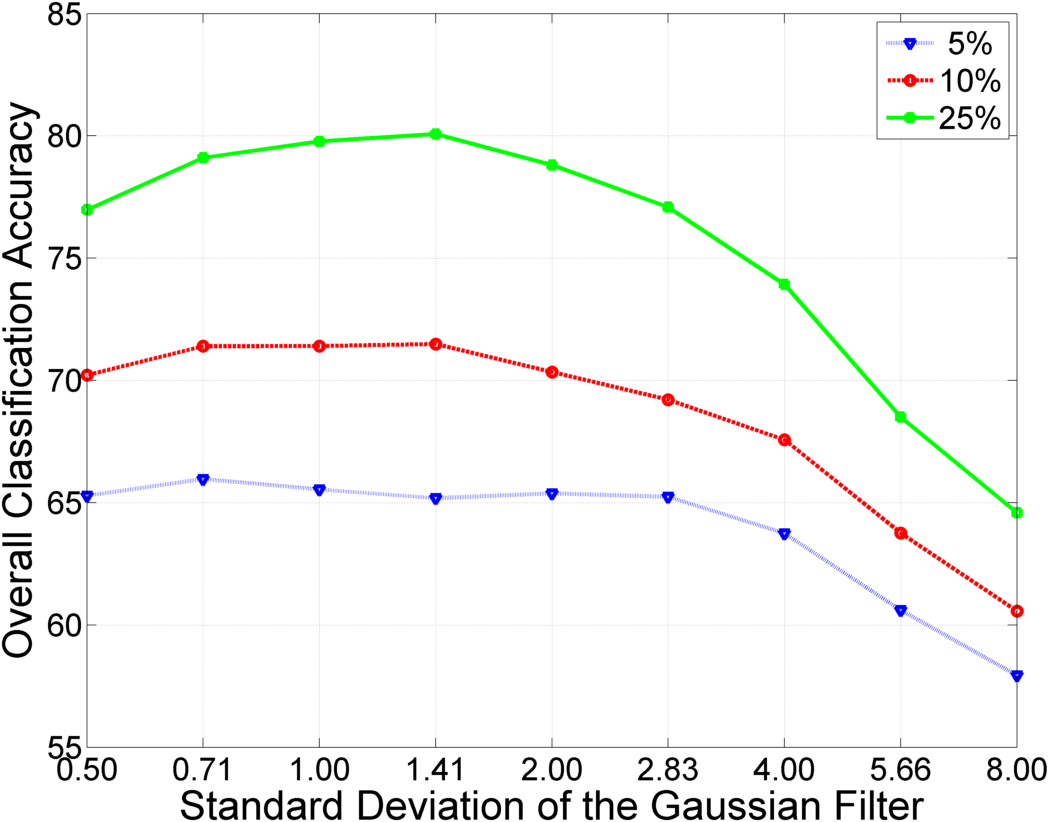}}
\subfigure[Indian Pines \& RF]{\includegraphics[width=0.23\textwidth, height = 0.2\textwidth]{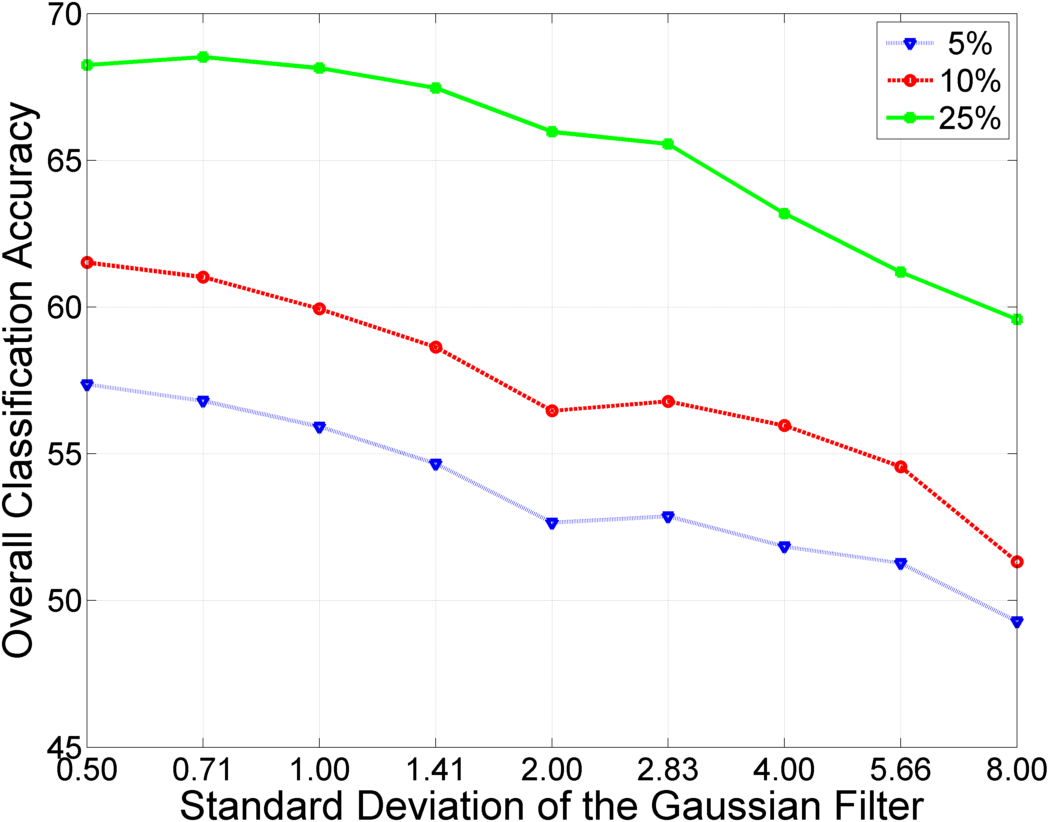}}
\subfigure[PaviaU \& SVM]{\includegraphics[width=0.23\textwidth, height = 0.2\textwidth]{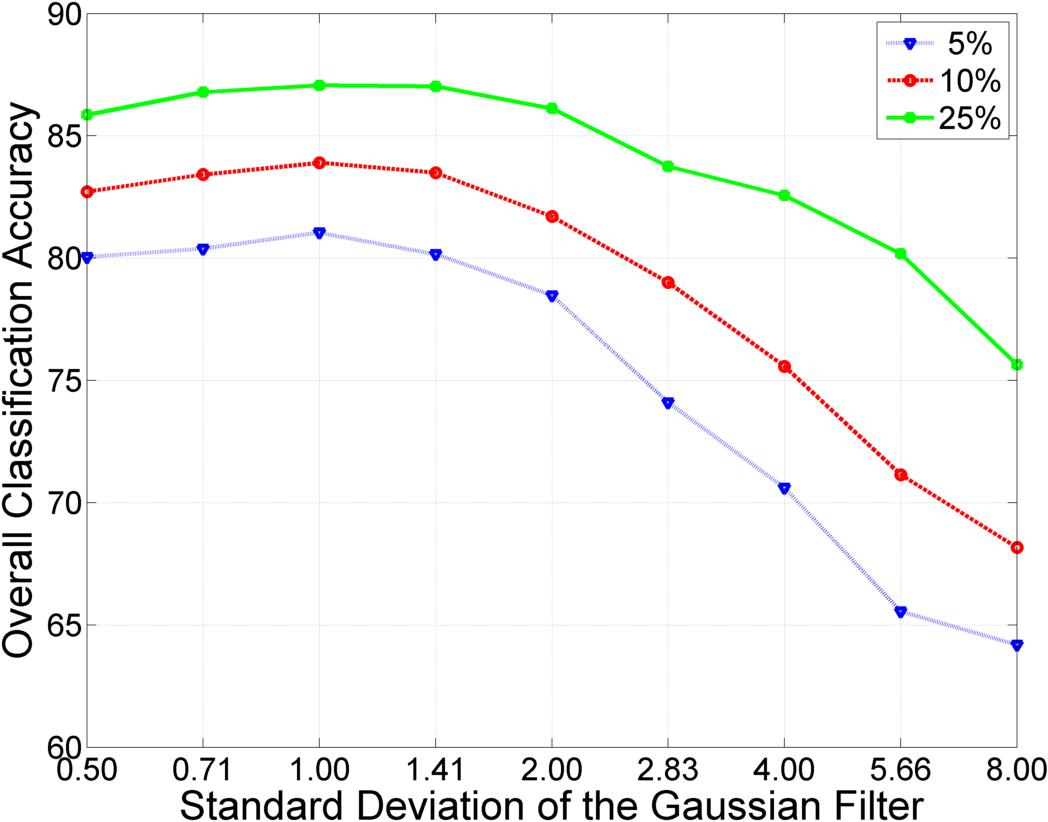}}
\subfigure[PaviaU \& RF]{\includegraphics[width=0.23\textwidth, height = 0.2\textwidth]{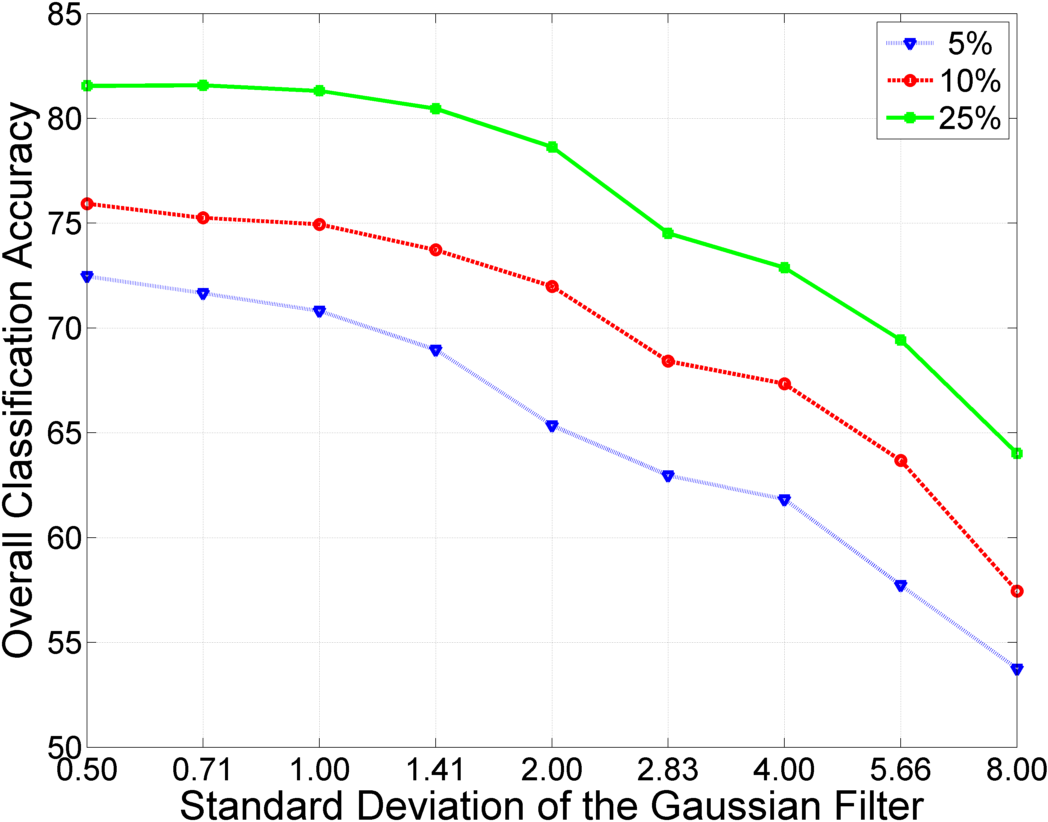}}
\caption{Classification accuracies vary with the standard deviation of Gaussian filter on the Indian Pines and Pavia University datasets under random sampling~(first row) and controlled random sampling~(second row) strategies.}
\label{fig:preprocessing2}
\end{center}
\end{figure}

We plot the classification accuracy as a function of the standard deviation in Fig.~\ref{fig:preprocessing2} for random sampling and controlled sampling method, respectively. From Fig.~\ref{fig:preprocessing2}~(a)-(d), we can see that the accuracy continuously increases until a specific point when Gaussian filter with larger standard deviation is used with random sampling strategy. This is consistent with the observation on the mean filter. We can assume that the Gaussian filter influences the data dependence to varying extents under different standard deviations, such that the classification accuracy is impacted by the filter parameter. This is also consistent with our earlier analysis that when data dependence is increased, the classification error bound will be tightened. However, this is not desired when evaluating a preprocessing method for image classification as we would like to know what is the actual contribution from the operation itself.

Compared with the random sampling, the controlled random sampling presents a different trend between the accuracy and standard deviation. The accuracy firstly improves marginally and then becomes stable or drops. This indicates that smoothing with an appropriate Gaussian operator can remove noises, and thus contribute to the final image classification. However, if the standard deviation of Gaussian is very large, too strong image smoothing does not help much for the discrimination of different classes since it may mix the training data with unlabelled data at boundaries of image regions, thus losing its adaptability. Under the new sampling strategy, Gaussian filter is able to improve the classification but not very significantly and the training and testing data dependence caused by overlap is no longer the dominant factor to the classification. Overall, these two experiments prove that the proposed sampling strategy is able to neutralize the improper benefit gained from enhancement of dependence between training and testing data.

\subsection{Raw Spectral Feature}
\begin{figure}[t]
\begin{center}
\subfigure[Training map]{\includegraphics[width=0.23\textwidth, height = 0.2\textwidth]{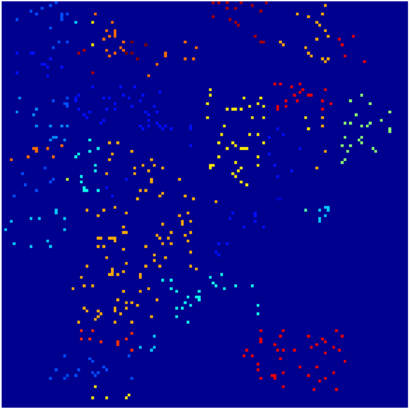}}
\subfigure[Classification map]{\includegraphics[width=0.23\textwidth, height = 0.2\textwidth]{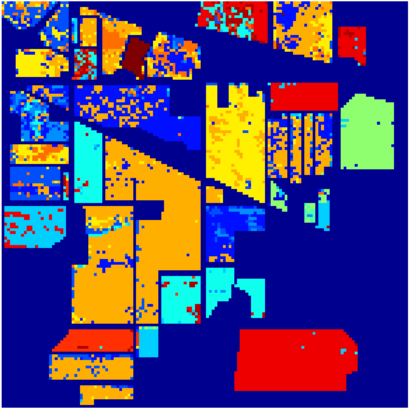}}
\subfigure[Training map]{\includegraphics[width=0.23\textwidth, height = 0.2\textwidth]{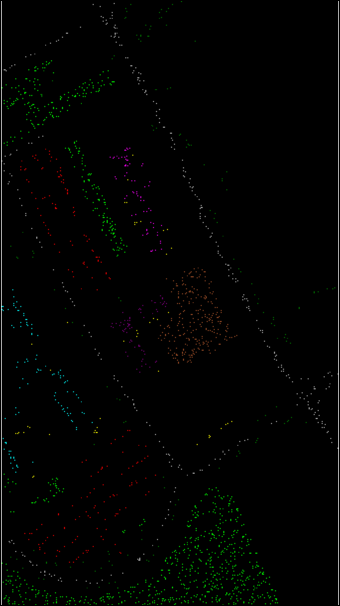}}
\subfigure[Classification map]{\includegraphics[width=0.23\textwidth, height = 0.2\textwidth]{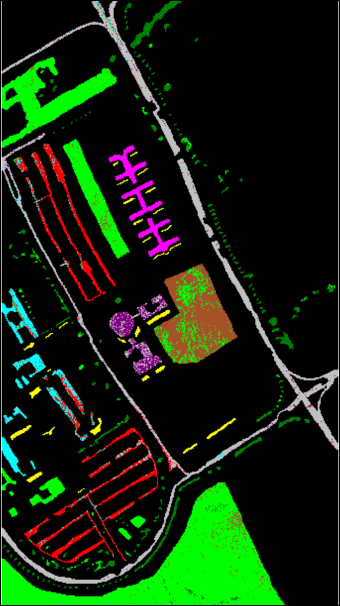}}
\\
\subfigure[Training map]{\includegraphics[width=0.23\textwidth, height = 0.2\textwidth]{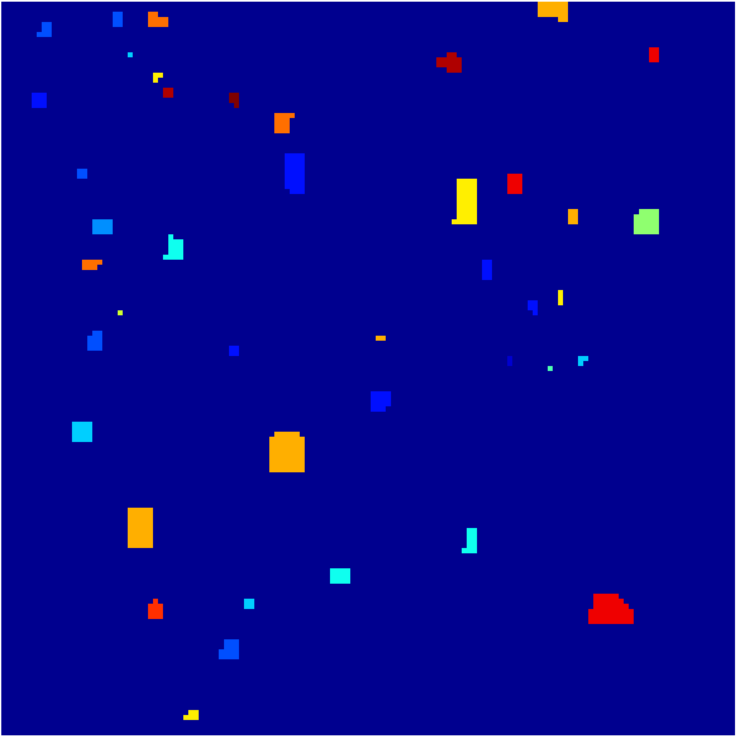}}
\subfigure[Classification map]{\includegraphics[width=0.23\textwidth, height = 0.2\textwidth]{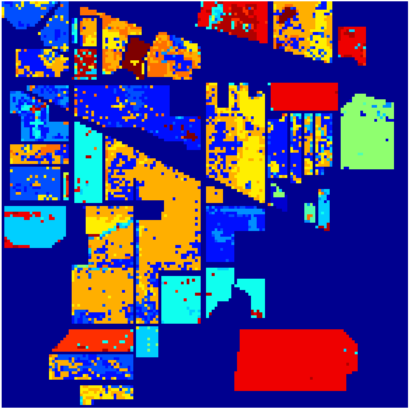}}
\subfigure[Training map]{\includegraphics[width=0.23\textwidth, height = 0.2\textwidth]{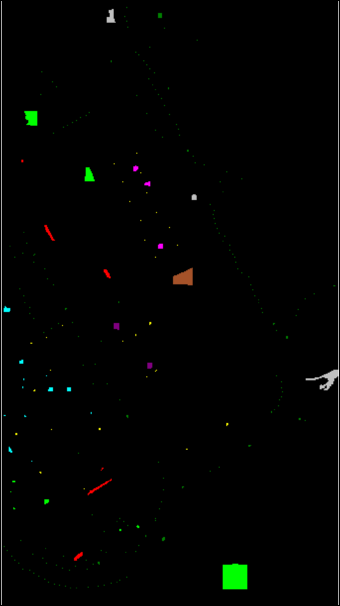}}
\subfigure[Classification map]{\includegraphics[width=0.23\textwidth, height = 0.2\textwidth]{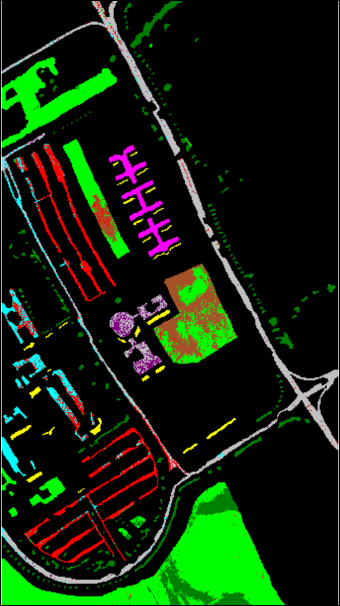}}
\caption{Training/classification maps on the Indian Pines and Pavia University datasets under random sampling~(first row) and controlled random sampling~(second row) strategies, when raw spectral features are used.}
\label{fig:rawmap}
\end{center}
\end{figure}

We then performed an experiment to compare two sampling strategies when raw spectral features were used on the benchmark datasets.
The objective of this experiment is to examine the effectiveness and objectiveness of the proposed sampling method compared to random sampling. As mentioned in Section~\ref{sec:intro}, there is no issue with the experimental setting with random sampling when evaluating a pixel based spectral feature. But we still do not know whether the proposed sampling method is qualified in such a task.

In the experiment, only the raw spectral features were used without any spatial processing. Other settings were same as the previous experiment such as the classifiers, repetition of experimental runs, etc. The overall accuracy and standard deviation under random sampling and the controlled random sampling strategies(*) are reported in Tables~\ref{tab:spectralspatial} and~\ref{tab:spectralspatial2} for SVM and RF, respectively. Following observations can be made from the results. Firstly, higher sampling rate leads to increase of classification accuracy on all datasets. This is the same and expected for both sampling methods. Secondly, the standard deviation of the accuracy from the proposed sampling strategy is much higher than that of the random courter part. This is due to the distinction of training data generated from the random seeds each time. Lastly, there is a reduction on the classification accuracy when the proposed sampling strategy is used. This is due to the fact that variations on the same class data in different regions are less sufficiently captured as some of them may not be included in the training samples when the proposed sampling strategy is used. The difference of accuracies is more evident on Indian Pines, Pavia University and Salinas datasets as these scenes include large blocks of regions in the same class, which leads to more benefits from spectral variation covered by random sampling strategy. For further illustrating this phenomenon, the classification maps on the Indian Pines and Pavia University under two sampling strategies are shown in Fig.~\ref{fig:rawmap}. Compared to random sampling, those testing samples far away from the training regions are easily misclassified under controlled random sampling.

Despite the differences, this does not affect a fair evaluation of different algorithms with the proposed sampling strategy. In this experiment, assuming that the goal is to evaluate SVM and RF, it can be concluded from the results that SVM is a preferred classifier since it generates higher classification accuracy. Therefore, although the new sampling strategy has made the hyperspectral classification a more challenging problem and forces more rigorous evaluation to the feature extraction and classification approaches, it is still qualified in evaluating the algorithms in hyperspectral image classification.

\subsection{Spectral-spatial Features}
Now we turn our attention to test the proposed sampling strategy with two typical spectral-spatial feature extraction methods, i.e., 3D discrete wavelet transform~(3D-DWT) and morphological profile. 3D-DWT is a typical example of filter based methods. The morphological profile is a widely adopted spatial feature extraction method, including a number of variations for hyperspectral image classification.

\subsubsection{3D discrete wavelet transform}
The discrete wavelet transform is derived from the wavelet transform which is a mathematical tool for signal analysis. Unlike Fourier transform, the advantage of wavelet transform is that the transformed signal provides time-frequency representation for the non-stationary signal, meaning that we can not only know whether a frequency component exists but also when it happens in a signal. The definition of continuous wavelet transform is shown as following:
\begin{equation}\label{eq:cwt}
\Psi_x^{\psi}(\tau,s)= \int x(t)\cdot\psi_{\tau,s}(t)dt
\end{equation}
where $\psi_{\tau,s}$ is the basis functions~(wavelet) with $s$ and $\tau$ that control the scale and translation, respectively.
When it comes to discrete samples, DWT is implemented by a series of filters in the frequency domain. Since hyperspectral images consist of three dimensions, 3D-DWT exploits the correlation along the wavelength axis, as well as along the spatial axes, so that both spatial and spectral structures of hyperspectral images can be more adequately mapped into the extracted features.

In the implementation, we followed the multiple scale setting as described in~\cite{Qian2013}, however, without the feature selection step. Firstly, the hyperspectral image was processed by a cascade of high pass filters and low pass filters. In each level, the data was decomposed into high frequency part and low frequency part. After three levels of decomposition, the original data was separated into 15 sub-cubes $C_1, C_2, ..., C_{15}$ based on the bandwidth, such that each of the sub-cubes contained different scales of information. To further capture the spatial distribution of hyperspectral images, a mean filter was applied on the sub-cubes:
\begin{equation}\label{eq:dwtmean}
\hat{C}_n(x,y,.) =  \frac{1}{9}\sum_{i=x-1}^{x+1}\sum_{j=y-1}^{y+1}C(x,y,.)
\end{equation}
In order to keep the sub-cube and the original data cube at the same size, the filtered signals were not down-sampled as what the traditional DWT does. Then these sub-cubes were concatenated into the wavelet features. The multidimensional function was carried out along two spatial dimensions $x$ and $y$, as well as the spectral dimension $\lambda$, respectively. The final concatenation worked as the feature for the whole data cube and can be represented as:
\begin{equation}\label{eq:dwt}
f(x,y) = (\hat{C}_{1}^{x},\hat{C}_{2}^{x},..., \hat{C}_{15}^{x}, \hat{C}_{1}^{y},\hat{C}_{2}^{y},..., \hat{C}_{15}^{y}, \hat{C}_{1}^{\lambda},\hat{C}_{2}^{\lambda},..., \hat{C}_{15}^{\lambda})
\end{equation}
where $f(x,y)$ is the 3D-DWT feature at location~(x,y).

The experimental results under random sampling strategy and controlled random sampling strategy(*) are shown in Table~\ref{tab:spectralspatial} and Table~\ref{tab:spectralspatial2} for SVM and RF, respectively.
As expected, controlled random sampling strategy leads to lower accuracy compared to random sampling strategy on all datasets.
Interesting observation can be obtained by comparing these results with the results on the raw spectral feature in Tables~\ref{tab:spectralspatial} and~\ref{tab:spectralspatial2}. On one hand, 3D-DWT performs better than raw spectral feature under both sampling methods. This indicates that the proposed method confirms that 3D-DWT is able to extract more discriminative information than raw spectral feature. On the other hand, under experimental setting with random sampling, 3D-DWT significantly improves the accuracy on all datasets over raw spectral feature. However, when testing it with the proposed controlled sampling strategy, the improvement can not reach the same level of significance, especially on Indian Pines, Pavia University, and Salinas datasets. It means that 3D-DWT does not perform that significantly better than the raw spectral features as expected, when eliminating the advantage of introducing information from the testing data into the training stage.

\subsubsection{Morphological profile}
To further analyse this issue, we undertook experiments on the mathematical morphology feature. Morphological operations employ the structuring elements in the image, making it possible to enhance or alleviate structures based on the specific requirements from users. The basic operators include erosion and dilation which expands and shrinks the structures, respectively. Combining them results in the opening~(erosion-dilation) and closing~(dilation-erosion) operations. These two processes can remove specific structures and noises without destroying the original primary structures in the image. The results of processing are called morphological profiles. Morphological profile based feature extraction method is able to explore the structures of objects based on the contrast and size of objects in the images, therefore, it has been widely studied for hyperspectral image classification~\cite{Benediktsson2005, DallaMura2011}.

We followed a basic implementation of extended morphological profiles~(EMP). The details of this method and its variation can be found in a survey paper from Fauvel et al~\cite{Fauvel2013}. The spatial feature was extracted as follows
\begin{equation}\label{eq:mm}
\Omega^{(n)}(I) = \left[o^{(n)}(I),...,o^{(1)}(I), I, c^{(1)}(I),...,c^{(n)}(I)\right]
\end{equation}
where $o^{(n)}(I)$ and $c^{(n)}(I)$ were the opening and closing operations with a disk-shape structural element of size $n$, respectively. As different sizes of structuring elements were used, the morphological profile $\Omega^{(n)}(I)$ was capable of integrating multi-scale information.
Before the feature extraction, a principle component analysis (PCA) step was applied to hyperspectral images to reduce the dimension of the data. Then the morphological profiles were obtained on each of the $m$ primary components:
\begin{equation}\label{eq:emp}
\hat{\Omega}^{(n)}_m(I) = \left[\Omega^{(n)}_1(I), \Omega^{(n)}_2(I), ...,\Omega^{(n)}_m(I)\right]
\end{equation}
In the last step, the morphological profiles were stacked with the spectral response to form the spectral-spatial feature.

The classification results with two sampling strategies are shown in Table~\ref{tab:spectralspatial} and Table~\ref{tab:spectralspatial2}. Similar to the results on 3D-DWT, although the morphological profile feature has achieved better performance than the raw spectral method when tested with random sampling strategy, the improvement is not as significant when controlled random sampling is used. This is mainly because that spectral-spatial method does not take much advantage of the overlapped information between training and testing samples under the proposed method.

Directly comparing two completely different spectral-spatial methods may not make much sense since different features are more suitable to extract features on specific datasets or sensitive to specific classifiers. Here we analyse the results from another point of view, which may explain the advantage of the proposed sampling over random sampling. In Table~\ref{tab:spectralspatial},
3D-DWT achieves higher accuracy than EMP on both Indian Pines and Pavia University datasets when random sampling is adopted. When adopting the new sampling strategy, 3D-DWT still performs slightly better than EMP on the Indian Pines, but EMP performs significantly better than 3D-DWT on the Pavia University.
This is consistent with the fact that the morphology method is capable of extracting more spatial structures than 3D-DWT on the dataset with high spatial resolution~\cite{Benediktsson2005}. Under the proposed sampling method, the properties of the spectral-spatial method can be more accurately reflected and evaluated in the experiments. This is impossible under random sampling because the classification result is strongly misled by the overlap between training and testing samples. Overall, the proposed sampling strategy reveals more real discriminative ability of the spectral-spatial methods, which is the purpose of the evaluation.

\section{Conclusion}
\label{sec:con}

This paper presented a comprehensive study on the influence of the widely adopted sampling strategy for performance evaluation of the spectral-spatial methods in hyperspectral image classification. We point out that random sampling has some problems because it has ignored the overlap and spatial dependency between training and testing samples when they are selected from the same image. Based on the \emph{non-i.i.d.} characteristic of hyperspectral image data, we proved that the improvement of classification accuracy by some spectral-spatial methods are partly due to the enhancement of dependence between training and testing data, compared with sole spectral information based methods. An alternative controlled random sampling strategy is proposed to alleviate these problems. This new strategy provides a better way to evaluate the effectiveness of spectral-spatial operations and the corresponding classifiers.

Finally, it should be noted that the aim of this paper is not to criticize the spectral-spatial methods themselves or the exploration of spatial information.
The concern is only on the widely adopted evaluation approach, or more strictly speaking, on the experimental setting.
Under the experimental setting with random sampling, the performance evaluation may be not equally fair and unbiased for all spectral-spatial methods.
This is especially the case for the practice that training and testing are performed on the same image. This problem is ultimately due to the lack of labelled hyperspectral data that are available for method evaluation. Therefore, a more urgent task for the research community is to build more benchmark datasets to facilitate future spectral-spatial hyperspectral image analysis research.

\begin{table*}[t]
\caption{Classification accuracies (overall accuracy and standard deviation) using raw spectral feature, 3D discrete wavelet transform~(3D-DWT) and extended morphological profile~(EMP) with random sampling and controlled random sampling(*), when linear SVM is adopted.}
\label{tab:spectralspatial}
\centering
\begin{tabular}{|m{1.5cm}*{9}{|>{\centering}p{1.0cm}}|} \hline
  \multirow{2}{*}{Dataset}&\multicolumn{3}{c|}{\textbf{Raw Spectral Feature}} & \multicolumn{3}{c|}{\textbf{3D-DWT}} & \multicolumn{3}{c|}{\textbf{EMP}} \tabularnewline \cline{2-10}
    & \%5      & \%10           & \%25               & \%5        & \%10           & \%25               & \%5        & \%10           & \%25 \tabularnewline \hline \hline
  Botswana     & $88.6\newline(1.4)$  &  $92.2\newline(1.1)$  &$95.0\newline(0.6)$  & $96.2\newline(0.7)$  &  $97.7\newline(0.5)$  &$99.4\newline(0.3)$  & $95.5\newline(2.0)$  &  $98.3\newline(0.6)$   &$99.5\newline(0.2)$ \tabularnewline \hline
  Botswana*    & $87.4\newline(1.4)$  &  $90.7\newline(0.5)$  &$93.0\newline(0.4)$ & $95.1\newline(1.4)$  &  $95.9\newline(1.3)$  &$96.6\newline(0.8)$  & $95.4\newline(1.2)$  &  $96.6\newline(0.7)$   &$97.6\newline(0.5)$ \tabularnewline \hline \hline
  Indian       & $72.5\newline(0.7)$  &  $77.1\newline(0.8)$  &$82.4\newline(0.4)$ & $88.1\newline(0.8)$  &  $93.7\newline(0.7)$  &$97.9\newline(0.3)$  &$83.0\newline(1.1)$  &  $88.2\newline(0.7)$   &$92.4\newline(0.4)$ \tabularnewline \hline
  Indian*      & $63.8\newline(2.2)$  &  $68.2\newline(1.7)$  &$75.0\newline(1.6)$ & $65.2\newline(3.0)$  &  $69.9\newline(3.0)$  &$79.1\newline(1.7)$ &$64.8\newline(2.8)$  &  $69.2\newline(2.7)$   &$77.2\newline(2.8)$ \tabularnewline \hline \hline
  KSC          & $76.1\newline(0.9)$  &  $80.4\newline(0.9)$  &$86.3\newline(0.7)$ & $87.7\newline(1.8)$  &  $91.8\newline(0.6)$  &$96.4\newline(0.5)$  &$76.9\newline(1.0)$  &  $83.3\newline(0.8)$   &$89.1\newline(0.5)$ \tabularnewline \hline
  KSC*         & $73.8\newline(2.1)$  &  $78.5\newline(1.1)$  &$83.8\newline(0.7)$ & $81.6\newline(2.2)$  &  $83.7\newline(2.6)$  &$87.9\newline(1.1)$  &$72.3\newline(3.6)$  &  $78.3\newline(2.7)$   &$84.5\newline(1.5)$ \tabularnewline \hline \hline
  PaviaU       & $89.9\newline(0.2)$  &  $90.7\newline(0.3)$  &$91.3\newline(0.2)$ & $97.8\newline(0.1)$  &  $98.6\newline(0.1)$  &$99.3\newline(0.1)$  &$97.0\newline(0.2)$  &  $97.6\newline(0.1)$   &$98.1\newline(0.1)$ \tabularnewline \hline
  PaviaU*      & $80.9\newline(3.9)$  &  $82.7\newline(4.0)$  &$84.5\newline(4.3)$ & $84.8\newline(3.2)$  &  $86.4\newline(3.4)$  &$89.2\newline(3.6)$  &$87.4\newline(3.1)$  &  $89.5\newline(1.7)$   &$91.7\newline(1.9)$ \tabularnewline \hline \hline
  Salinas      & $92.4\newline(0.1)$  &  $92.8\newline(0.1)$  &$93.1\newline(0.1)$ & $96.4\newline(0.1)$  &  $97.3\newline(0.1)$  &$98.3\newline(0.1)$  &$94.5\newline(0.3)$  &  $95.0\newline(0.2)$   &$95.3\newline(0.1)$ \tabularnewline \hline
  Salinas*     & $81.8\newline(2.7)$  &  $81.6\newline(3.8)$  &$83.0\newline(3.3)$ & $80.9\newline(3.0)$  &  $82.2\newline(3.6)$  &$83.4\newline(3.4)$  &$83.5\newline(1.4)$  &  $85.0\newline(2.4)$   &$84.8\newline(2.4)$ \tabularnewline
  \hline
\end{tabular}
\end{table*}
\begin{table*}[t]
\caption{Classification accuracies (overall accuracy and standard deviation) using raw spectral feature, 3D discrete wavelet transform~(3D-DWT) and extended morphological profile~(EMP) with random sampling and controlled random sampling(*), when RF is adopted.}
\label{tab:spectralspatial2}
\centering
\begin{tabular}{|m{1.5cm}*{9}{|>{\centering}p{1.0cm}}|} \hline
  \multirow{2}{*}{Dataset}&\multicolumn{3}{c|}{\textbf{Raw Spectral Feature}} & \multicolumn{3}{c|}{\textbf{3D-DWT}} & \multicolumn{3}{c|}{\textbf{EMP}} \tabularnewline \cline{2-10}
    & \%5      & \%10           & \%25               & \%5        & \%10           & \%25               & \%5        & \%10           & \%25 \tabularnewline \hline \hline
  Botswana     & $82.4\newline(1.6)$  &  $85.4\newline(0.8)$  &$88.7\newline(0.5)$  & $90.5\newline(1.8)$  &  $94.3\newline(1.1)$  &$97.7\newline(0.5)$  & $90.4\newline(1.6)$  &  $94.2\newline(1.0)$   &$97.3\newline(0.9)$ \tabularnewline \hline
  Botswana*    & $80.5\newline(1.6)$  &  $82.8\newline(2.0)$  &$85.9\newline(1.4)$ & $88.3\newline(1.5)$  &  $90.7\newline(1.4)$  &$93.5\newline(1.1)$  & $88.4\newline(1.6)$  &  $91.6\newline(1.2)$   &$94.6\newline(1.1)$ \tabularnewline \hline \hline
  Indian       & $70.5\newline(0.9)$  &  $75.6\newline(0.9)$  &$81.4\newline(0.5)$ & $75.1\newline(1.4)$  &  $81.6\newline(0.8)$  &$89.7\newline(0.5)$  &$81.1\newline(1.5)$  &  $88.0\newline(0.9)$   &$93.7\newline(0.6)$ \tabularnewline \hline
  Indian*      & $56.7\newline(1.5)$  &  $61.3\newline(2.5)$  &$66.7\newline(2.9)$ & $57.4\newline(2.1)$  &  $61.4\newline(2.1)$  &$67.5\newline(1.6)$ &$64.4\newline(1.5)$  &  $69.6\newline(2.7)$   &$76.2\newline(3.0)$ \tabularnewline \hline \hline
  KSC          & $82.9\newline(0.6)$  &  $86.6\newline(0.8)$  &$90.1\newline(0.3)$ & $82.2\newline(1.3)$  &  $88.7\newline(0.7)$  &$92.9\newline(0.6)$  &$87.1\newline(1.0)$  &  $91.8\newline(1.1)$   &$95.5\newline(0.5)$ \tabularnewline \hline
  KSC*         & $77.1\newline(2.4)$  &  $79.8\newline(1.7)$  &$84.5\newline(1.7)$ & $74.8\newline(2.2)$  &  $79.1\newline(2.1)$  &$84.6\newline(1.6)$  &$80.6\newline(2.8)$  &  $84.4\newline(1.5)$   &$89.8\newline(1.8)$ \tabularnewline \hline \hline
  PaviaU       & $87.3\newline(0.4)$  &  $89.3\newline(0.2)$  &$91.4\newline(0.1)$ & $92.4\newline(0.3)$  &  $94.1\newline(0.2)$  &$96.1\newline(0.1)$  &$95.5\newline(0.2)$  &  $97.1\newline(0.2)$   &$98.5\newline(0.1)$ \tabularnewline \hline
  PaviaU*      & $71.2\newline(5.0)$  &  $73.6\newline(4.3)$  &$81.4\newline(2.1)$ & $75.8\newline(3.6)$  &  $79.0\newline(3.2)$  &$83.4\newline(2.2)$  &$78.2\newline(5.2)$  &  $80.9\newline(4.5)$   &$88.2\newline(2.3)$ \tabularnewline \hline \hline
  Salinas      & $90.3\newline(0.1)$  &  $91.5\newline(0.2)$  &$93.0\newline(0.1)$ & $93.0\newline(0.3)$  &  $94.2\newline(0.2)$  &$95.7\newline(0.1)$  &$94.9\newline(0.3)$  &  $96.4\newline(0.2)$   &$97.7\newline(0.2)$ \tabularnewline \hline
  Salinas*     & $79.0\newline(2.6)$  &  $80.9\newline(3.9)$  &$84.1\newline(2.6)$ & $77.6\newline(2.9)$  &  $80.6\newline(2.6)$  &$83.8\newline(1.8)$  &$82.0\newline(1.6)$  &  $84.4\newline(2.8)$   &$86.6\newline(2.6)$ \tabularnewline
  \hline
\end{tabular}
\end{table*}
\clearpage

\renewcommand{\refname}{REFERENCE} 
\bibliographystyle{myIEEEtran}
\bibliography{refs}

\begin{thebibliography}{10}
\providecommand{\url}[1]{#1}
\csname url@samestyle\endcsname
\providecommand{\newblock}{\relax}
\providecommand{\bibinfo}[2]{#2}
\providecommand{\BIBentrySTDinterwordspacing}{\spaceskip=0pt\relax}
\providecommand{\BIBentryALTinterwordstretchfactor}{4}
\providecommand{\BIBentryALTinterwordspacing}{\spaceskip=\fontdimen2\font plus
\BIBentryALTinterwordstretchfactor\fontdimen3\font minus
  \fontdimen4\font\relax}
\providecommand{\BIBforeignlanguage}[2]{{%
\expandafter\ifx\csname l@#1\endcsname\relax
\typeout{** WARNING: IEEEtran.bst: No hyphenation pattern has been}%
\typeout{** loaded for the language `#1'. Using the pattern for}%
\typeout{** the default language instead.}%
\else
\language=\csname l@#1\endcsname
\fi
#2}}
\providecommand{\BIBdecl}{\relax}
\BIBdecl

\bibitem{Lu2007Survey}
D.~Lu and Q.~Weng, ``A survey of image classification methods and techniques
  for improving classification performance,'' \emph{International Journal of
  Remote Sensing}, vol.~28, no.~5, pp. 823--870, Jan. 2007.

\bibitem{Chen2011}
Y.~Chen, N.~Nasrabadi, and T.~Tran, ``Hyperspectral image classification using
  dictionary-based sparse representation,'' \emph{IEEE Transactions on
  Geoscience and Remote Sensing}, vol.~49, no.~10, pp. 3973--3985, Oct 2011.

\bibitem{Fauvel2013}
M.~Fauvel, Y.~Tarabalk, J.~Benediktsson, J.~Chanussot, and J.~Tilton,
  ``Advances in spectral–spatial classification of hyperspectral images,''
  \emph{Proceedings of the IEEE}, vol. 101, no.~3, pp. 652--675, 2013.

\bibitem{Moser2013Markov}
G.~Moser, S.~Serpico, and J.~Benediktsson, ``Land-cover mapping by {Markov}
  modeling of spatial-contextual information in very-high-resolution remote
  sensing images,'' \emph{Proceedings of the IEEE}, vol. 101, no.~3, pp.
  631--651, March 2013.

\bibitem{Friedl2000}
M.~A. Friedl, C.~Woodcock, S.~Gopal, D.~Muchoney, A.~H. Strahler, and
  C.~Barker-Schaaf, ``A note on procedures used for accuracy assessment in land
  cover maps derived from {AVHRR} data,'' \emph{International Journal of Remote
  Sensing}, vol.~21, no.~5, pp. 1073--1077, 2000.

\bibitem{Stehman1998Assessment}
\BIBentryALTinterwordspacing
S.~V. Stehman and R.~L. Czaplewski, ``Design and analysis for thematic map
  accuracy assessment: Fundamental principles,'' \emph{Remote Sensing of
  Environment}, vol.~64, no.~3, pp. 331 -- 34, 1998.
\BIBentrySTDinterwordspacing

\bibitem{Richards1999}
J.~A. Richards and X.~Jia, \emph{Remote Sensing Digital Image Analysis: An
  Introduction}, 3rd~ed., D.~E. Ricken and W.~Gessner, Eds.\hskip 1em plus
  0.5em minus 0.4em\relax Secaucus, NJ, USA: Springer-Verlag New York, Inc.,
  1999.

\bibitem{Foody2002Assessment}
G.~Foody, ``Status of land cover classification accuracy assessment,''
  \emph{Remote Sensing of Environment}, vol.~80, no.~1, pp. 185--201, 2002.

\bibitem{Li2012}
J.~Li, J.~Bioucas-Dias, and A.~Plaza, ``Spectral–spatial hyperspectral image
  segmentation using subspace multinomial logistic regression and {Markov}
  random fields,'' \emph{IEEE Transactions on Geoscience and Remote Sensing},
  vol.~50, no.~3, pp. 809--823, 2012.

\bibitem{Qian2013}
Y.~Qian, M.~Ye, and J.~Zhou, ``Hyperspectral image classification based on
  structured sparse logistic regression and three-dimensional wavelet texture
  features,'' \emph{IEEE Transactions on Geoscience and Remote Sensing},
  vol.~51, no.~4, pp. 2276--2291, 2013.

\bibitem{Fang2014}
L.~Fang, S.~Li, X.~Kang, and J.~Benediktsson, ``Spectral–spatial
  hyperspectral image classification via multiscale adaptive sparse
  representation,'' \emph{IEEE Transactions on Geoscience and Remote Sensing},
  vol.~52, no.~12, pp. 7738--7749, 2014.

\bibitem{Zhen2013}
Z.~Zhen, L.~J. Quackenbush, S.~V. Stehman, and L.~Zhang, ``Impact of training
  and validation sample selection on classification accuracy and accuracy
  assessment when using reference polygons in object-based classification,''
  \emph{International Journal of Remote Sensing}, vol.~34, no.~19, pp.
  6914--6930, 2013.

\bibitem{Ye2015}
M.~Ye, Y.~Qian, and J.~Zhou, ``Multi-task sparse nonnegative matrix
  factorization for joint spectral-spatial hyperspectral imagery denoising,''
  \emph{IEEE Transactions on Geoscience and Remote Sensing}, vol.~53, no.~5,
  pp. 2621--2639, 2015.

\bibitem{Velasco2010}
S.~Velasco-Forero and J.~Angulo, ``Spatial structures detection in
  hyperspectral images using mathematical morphology,'' in \emph{2nd Workshop
  on Hyperspectral Image and Signal Processing: Evolution in Remote Sensing
  (WHISPERS)}, 2010, pp. 1--4.

\bibitem{Tarabalka2010}
Y.~Tarabalka, J.~Chanussot, and J.~A. Benediktsson, ``Spectral–spatial
  hyperspectral image segmentation using subspace multinomial logistic
  regression and {Markov} random fields,'' \emph{Pattern Recognition}, vol.~43,
  no.~7, pp. 2367--2379, 2010.

\bibitem{Jia2015}
S.~Jia, L.~Shen, and Q.~Li, ``Gabor feature-based collaborative representation
  for hyperspectral imagery classification,'' \emph{IEEE Transactions on
  Geoscience and Remote Sensing}, vol.~53, no.~2, pp. 1118--1129, 2015.

\bibitem{Tang2015}
Y.~Tang, Y.~Lu, and H.~Yuan, ``Hyperspectral image classification based on
  three-dimensional scattering wavelet transform,'' \emph{IEEE Transactions on
  Geoscience and Remote Sensing}, vol.~53, no.~5, pp. 2467--2480, 2015.

\bibitem{WLi2015}
W.~Li, C.~Chen, H.~Su, and Q.~Du, ``Local binary patterns and extreme learning
  machine for hyperspectral imagery classification,'' \emph{IEEE Transactions
  on Geoscience and Remote Sensing}, vol.~53, no.~7, pp. 3681--3693, 2015.

\bibitem{Benediktsson2005}
J.~Benediktsson, J.~Palmason, and J.~Sveinsson, ``Classification of
  hyperspectral data from urban areas based on extended morphological
  profiles,'' \emph{IEEE Transactions on Geoscience and Remote Sensing},
  vol.~43, no.~3, pp. 480--491, 2005.

\bibitem{Fauvel2008}
M.~Fauvel, J.~A. Benediktsson, J.~Chanussot, and J.~R. Sveinsson, ``{Spectral
  and Spatial Classification of Hyperspectral Data Using SVMs and Morphological
  Profiles},'' \emph{IEEE Transactions on Geoscience and Remote Sensing},
  vol.~46, no.~11, pp. 3804--3814, 2008.

\bibitem{DallaMura2011}
M.~{Dalla Mura}, A.~Villa, J.~A. Benediktsson, J.~Chanussot, and L.~Bruzzone,
  ``{Classification of Hyperspectral Images by Using Extended Morphological
  Attribute Profiles and Independent Component Analysis},'' \emph{IEEE
  Geoscience and Remote Sensing Letters}, vol.~8, no.~3, pp. 542--546, 2011.

\bibitem{Liang2013}
J.~Liang, J.~Zhou, X.~Bai, and Y.~Qian, ``Salient object detection in
  hyperspectral imagery,'' in \emph{20th IEEE International Conference on Image
  Processing}, 2013, pp. 2393--2397.

\bibitem{Nina2010}
F.~Nina-Paravecino and V.~Manian, ``Spherical harmonics as a shape descriptor
  for hyperspectral image classification,'' in \emph{Proc. SPIE 7695,
  Algorithms and Technologies for Multispectral, Hyperspectral, and
  Ultraspectral Imagery XVI}, 2010, p. 76951E.

\bibitem{Khuwuthyakorn2011}
P.~Khuwuthyakorn, A.~Robles-Kelly, and J.~Zhou, ``Affine invariant
  hyperspectral image descriptors based upon harmonic analysis,'' in
  \emph{Machine Vision Beyond the Visible Spectrum}, R.~Hammoud, G.~Fan,
  R.~McMillan, and K.~Ikeuchi, Eds.\hskip 1em plus 0.5em minus 0.4em\relax
  Springer, 2011.

\bibitem{Jia2013}
X.~Jia, B.~Kuo, and M.~Crawford, ``Feature mining for hyperspectral image
  classification,'' \emph{Proceedings of the IEEE}, vol. 101, no.~3, pp.
  676--697, 2013.

\bibitem{Pu2014}
H.~Pu, Z.~Chen, B.~Wang, and G.~Jiang, ``A novel spatial–spectral similarity
  measure for dimensionality reduction and classification of hyperspectral
  imagery,'' \emph{IEEE Transactions on Geoscience and Remote Sensing},
  vol.~52, no.~11, pp. 7008--7022, 2014.

\bibitem{Tarabalka2010letters}
Y.~Tarabalka, M.~Fauvel, J.~Chanussot, and J.~A. Benediktsson, ``{SVM}- and
  {MRF}-based method for accurate classification of hyperspectral images,''
  \emph{IEEE Geoscience and Remote Sensing Letters}, vol.~7, no.~4, pp.
  736--740, 2010.

\bibitem{Sun2015}
L.~Sun, Z.~Wu, J.~Liu, L.~Xiao, and Z.~Wei, ``Supervised spectral–spatial
  hyperspectral image classification with weighted {Markov} random fields,''
  \emph{IEEE Geoscience and Remote Sensing Letters}, vol.~53, no.~3, pp.
  1490--1503, 2015.

\bibitem{Zhong2010}
P.~Zhong and R.~Wang, ``Learning conditional random fields for classification
  of hyperspectral images,'' \emph{IEEE Transactions on Image Processing},
  vol.~19, no.~7, p. 1890–1907, 2010.

\bibitem{Ji2014}
R.~Ji, Y.~Gao, R.~Hong, Q.~Liu, D.~Tao, and X.~Li, ``Spectral-spatial
  constraint hyperspectral image classification,'' \emph{IEEE Transactions on
  Geoscience and Remote Sensing}, vol.~52, no.~3, pp. 1811--1824, 2014.

\bibitem{Camps-Valls2006}
\BIBentryALTinterwordspacing
G.~Camps-Valls, L.~Gomez-Chova, J.~Munoz-Mari, J.~Vila-Frances, and
  J.~Calpe-Maravilla, ``{Composite Kernels for Hyperspectral Image
  Classification},'' \emph{IEEE Geoscience and Remote Sensing Letters}, vol.~3,
  no.~1, pp. 93--97, jan 2006.
\BIBentrySTDinterwordspacing

\bibitem{Dias2013}
J.~Bioucas-Dias, A.~Plaza, G.~Camps-Valls, P.~Scheunders, N.~Nasrabadi, and
  J.~Chanussot, ``Hyperspectral remote sensing data analysis and future
  challenges,'' \emph{IEEE Geoscience and Remote Sensing Magazine}, vol.~1,
  no.~2, pp. 6--36, 2013.

\bibitem{1993}
\BIBentryALTinterwordspacing
P.~Legendre, ``\BIBforeignlanguage{English}{Spatial autocorrelation: Trouble or
  new paradigm?}'' \emph{\BIBforeignlanguage{English}{Ecology}}, vol.~74,
  no.~6, pp. pp. 1659--1673, 1993.
\BIBentrySTDinterwordspacing

\bibitem{Li2013}
J.~Li, J.~Bioucas-Dias, and A.~Plaza, ``Spectral–spatial classification of
  hyperspectral data using loopy belief propagation and active learning,''
  \emph{IEEE Transactions on Geoscience and Remote Sensing}, vol.~51, no.~2,
  pp. 844--856, 2013.

\bibitem{Wang2014}
L.~Wang, S.~Hao, Y.~Wang, Y.~Lin, and Q.~Wang, ``Spatial–spectral
  information-based semisupervised classification algorithm for hyperspectral
  imagery,'' \emph{IEEE Journal of Selected Topics in Applied Earth
  Observations and Remote Sensing}, vol.~7, no.~8, pp. 3577--3585, 2014.

\bibitem{datasets}
``Hyperspectral remote sensing scenes,''
  \url{http://www.ehu.eus/ccwintco/index.php?title=Hyperspectral_Remote_Sensing_Scenes},
  accessed: 2015-01-28.

\bibitem{Mohri2012}
M.~Mohri, A.~Rostamizadeh, and A.~Talwalkar, \emph{{Foundations of Machine
  Learning}}.\hskip 1em plus 0.5em minus 0.4em\relax MIT Press, 2012.

\bibitem{Bartlett2002}
\BIBentryALTinterwordspacing
P.~P.~L. Bartlett and S.~Mendelson, ``{Rademacher and {Gaussian} Complexities:
  Risk Bounds and Structural Results},'' \emph{Journal of Machine Learning
  Research}, vol.~3, no.~3, pp. 463--482, 2002.
\BIBentrySTDinterwordspacing

\bibitem{McDonald2011}
\BIBentryALTinterwordspacing
D.~J. McDonald, C.~R. Shalizi, and M.~Schervish, ``{Estimating beta-mixing
  coefficients},'' p.~9, mar 2011.
\BIBentrySTDinterwordspacing

\bibitem{Yu1994}
\BIBentryALTinterwordspacing
B.~Yu, ``{Rates of convergence for empirical processes of stationary mixing
  sequences},'' \emph{The Annals of Probability}, 1994.
\BIBentrySTDinterwordspacing

\bibitem{Vidyasagar2013}
\BIBentryALTinterwordspacing
M.~Vidyasagar, \emph{{Learning and Generalisation: With Applications to Neural
  Networks}}.\hskip 1em plus 0.5em minus 0.4em\relax Springer Science {\&}
  Business Media, 2013.
\BIBentrySTDinterwordspacing

\bibitem{Mohri2008}
\BIBentryALTinterwordspacing
M.~Mohri and A.~Rostamizadeh, ``{Rademacher Complexity Bounds for Non-{I.I.D.}
  Processes},'' in \emph{Advances in Neural Information Processing Systems 21},
  2008, pp. 1097--1104.
\BIBentrySTDinterwordspacing

\bibitem{Adams1994}
\BIBentryALTinterwordspacing
R.~Adams and L.~Bischof, ``{Seeded region growing},'' \emph{IEEE Transactions
  on Pattern Analysis and Machine Intelligence}, vol.~16, no.~6, pp. 641--647,
  jun 1994.
\BIBentrySTDinterwordspacing

\bibitem{Gislason2006}
\BIBentryALTinterwordspacing
P.~O. Gislason, J.~A. Benediktsson, and J.~R. Sveinsson, ``{Random Forests for
  land cover classification},'' \emph{Pattern Recognition Letters}, vol.~27,
  no.~4, pp. 294--300, mar 2006.
\BIBentrySTDinterwordspacing

\bibitem{Chang2011}
C.-C. Chang and C.-J. Lin, ``{LIBSVM}: A library for support vector machines,''
  \emph{ACM Transactions on Intelligent Systems and Technology}, vol.~2, pp.
  27:1--27:27, 2011, software available at
  \url{http://www.csie.ntu.edu.tw/~cjlin/libsvm}.

\bibitem{Hall2009}
\BIBentryALTinterwordspacing
M.~Hall, E.~Frank, G.~Holmes, B.~Pfahringer, P.~Reutemann, and I.~H. Witten,
  ``The weka data mining software: An update,'' \emph{SIGKDD Explor. Newsl.},
  vol.~11, no.~1, pp. 10--18, Nov. 2009.
\BIBentrySTDinterwordspacing

\end{thebibliography}

\end{document}